\documentclass[11pt, english]{article}

\usepackage{lmodern}
\usepackage[T1]{fontenc}
\usepackage[utf8]{inputenc}
\usepackage[a4paper, margin=1in]{geometry}

\usepackage{amsmath}
\usepackage{amssymb}
\usepackage{amsfonts}
\usepackage{siunitx}

\usepackage{graphicx}
\usepackage{xcolor}

\usepackage{booktabs}
\usepackage{tabularx}
\usepackage{longtable}
\usepackage{array}
\usepackage[ruled,linesnumbered]{algorithm2e}

\usepackage{multicol}
\usepackage{multirow}
\usepackage{caption}
\usepackage{subcaption}

\usepackage{hyperref}
\hypersetup{
    colorlinks=true,
    linkcolor=blue,
    citecolor=blue,
    urlcolor=blue,
    pdftitle={Beyond Backpropagation: Exploring Innovative Algorithms for Energy-Efficient Deep Neural Network Training},
    pdfauthor={Przemyslaw Spyra}
}

\title{Beyond Backpropagation: Exploring Innovative Algorithms for Energy-Efficient Deep Neural Network Training}

\author{
    Przemyslaw Spyra \\
    \texttt{przspyra@student.agh.edu.pl} \\\\
    AGH University of Science and Technology \\
    Krakow, Poland
}

\date{}

\newcommand{\code}[1]{\texttt{#1}}

\begin{document}

\maketitle

\begin{abstract}
The increasing computational demands of deep neural networks (DNNs), primarily driven by the high-energy requirements of the backpropagation (BP) algorithm, pose a significant challenge to sustainable AI development. This paper presents a rigorous investigation into three promising training methods without backpropagation: the \textbf{Forward-Forward (FF) Algorithm}, the \textbf{Cascaded-Forward (CaFo) Algorithm}, and the \textbf{Mono-Forward (MF) Algorithm}. This investigation charts an evolutionary progression from foundational concepts to a demonstrably superior solution.

To ensure a fair and insightful analysis, a robust comparative framework was established. Each alternative algorithm was implemented on its native architecture (MLPs for FF and MF, a CNN for CaFo) and systematically compared against an equivalent model trained with BP. Critically, this methodology mandates comprehensive hyperparameter optimization for all algorithms using Optuna, coupled with consistent early stopping criteria based on validation performance, thereby ensuring that all comparisons are conducted between optimally tuned models.

The investigation culminates in a significant finding: The Mono-Forward (MF) algorithm not only competes with, but consistently surpasses backpropagation in classification accuracy on its native MLP architectures. This superior generalization arises from the algorithm's ability to converge to a more favorable minimum in the validation loss landscape, challenging the long-held assumption that global optimization is a prerequisite for achieving state-of-the-art performance. This breakthrough delivers profound efficiency gains, reducing energy consumption by up to 41\% and accelerating training by up to 34\% on complex datasets. These figures represent direct measurements at the hardware level obtained via the NVIDIA Management Library (NVML) API, which translate to a demonstrably smaller carbon footprint as estimated by the CodeCarbon tool.

Beyond this primary result, this paper provides a detailed analysis of the underlying drivers of computational efficiency. This "autopsy at the hardware level" reveals the sources of performance disparities by exposing the architectural inefficiencies of the FF algorithm while validating the computationally lean design of MF. Furthermore, this research empirically challenges the simplistic notion that all BP-free methods are inherently more memory-efficient, instead revealing the nuanced resource requirements of each approach. By documenting the trajectory from FF's foundational concepts and their practical limitations to MF's compelling synthesis of superior accuracy and sustainability, this work offers a clear, data-driven roadmap for the future of energy-efficient deep learning.
\end{abstract}

%
%
%
\section{Introduction}
\label{chap:introduction}

\subsection{Motivation}
\label{sec:motivation}

Deep neural networks (DNNs) have become a transformative technology in machine learning and artificial intelligence, establishing state-of-the-art benchmarks in diverse domains such as computer vision \cite{krizhevsky2012imagenet, he2016deep}, natural language processing \cite{vaswani2017attention, devlin2018bert}, and reinforcement learning \cite{mnih2015human}. However, this success is predicated on a substantial and escalating demand for computational resources, which translates directly into significant energy consumption. The challenge is particularly acute, as contemporary models now routinely scale to billions or trillions of parameters. The predominant training paradigm, which relies on the backpropagation (BP) algorithm \cite{rumelhart1986learning}, which requires the storage of all intermediate activations during the forward-pass to facilitate gradient computation in the subsequent backward pass. This requirement results in considerable memory usage and a correspondingly high energy budget \cite{chen2016training}. Moreover, the inherently sequential nature of the forward and backward passes imposes a fundamental constraint on parallelization, a bottleneck known as "backward locking" \cite{jaderberg2017decoupled}.

The environmental consequences of these computational demands have emerged as a pressing concern. Recent analyses reveal a paradox in modern AI, where its capacity to optimize energy systems is contrasted with its own operational requirements. Although specific projections vary according to scope and underlying assumptions, some forecast that AI could account for \textbf{up to 3\% of global electricity demand by 2030} \cite{wef2025paradox}, underscoring an undisputed trend of rapid escalating energy consumption. Architectural studies confirm that even modest reductions in network complexity can halve the energy consumed during training \cite{yarally2023uncovering}, while hardware-aware analyzes underscore the non-linear relationship between model scale and energy cost \cite{desislavov2024energy}. In particular, training a single large language model can produce a carbon footprint of \textbf{more than 500 metric tons of CO2}, a figure equivalent to the lifetime emissions of several conventional automobiles \cite{patterson2021carbon, abbas2024impact}. In response, a new generation of hardware accelerators, sparsity techniques, and analytical tools such as CodeCarbon \cite{codecarbon2021}, which provides an estimated CO2e from energy use and local grid intensity, is being developed to tackle this energy crisis.

This context has spurred a search for alternative training algorithms that completely eliminate the backward pass. Among the most promising are Geoffrey Hinton's Forward-Forward (FF) algorithm \cite{hinton2022forward}, the Cascaded-Forward (CaFo) algorithm \cite{zhao2023cafo}, and the Mono-Forward (MF) algorithm \cite{gong2025mono}. These backpropagation-free (BP-free) methods are not only intended to reduce energy consumption but are also inspired by learning principles observed in biological neural systems. In contrast to backpropagation's dependence on a global error signal and non-local computations, these alternatives employ local, layer-wise learning mechanisms. Such mechanisms are analogous to synaptic plasticity in the brain, where learning is governed by localized phenomena such as Hebbian learning \cite{song2000competitive} and spike-timing dependent plasticity (STDP) \cite{bi1998synaptic}.

By aligning more closely with these neurobiological principles, these alternative algorithms offer potential improvements in energy efficiency and scalability. They can reduce computational overhead and memory requirements and, by breaking the global dependency of the backward pass, create new opportunities for parallel and distributed processing across diverse architectures, including multi-layer perceptrons (MLPs) and convolutional neural networks (CNNs). This localized learning paradigm could be especially advantageous in resource-constrained environments, such as on edge devices or within Internet of Things (IoT) systems. Ultimately, the biological inspiration that underpins these methods fosters the potential to develop more robust and efficient learning strategies, bridging the gap between artificial and natural intelligence and paving the way for a new generation of deep learning algorithms that are powerful and sustainable \cite{bengio2015towards, lillicrap2020backpropagation}.

\subsection{Problem Statement}
\label{sec:problem}

The conventional backpropagation (BP) algorithm, despite its widespread success, is constrained by inherent limitations. Currently, the emerging class of BP-free alternatives introduces its own set of unresolved challenges related to practical viability and comparative performance. First, standard BP is constrained by significant drawbacks, including high memory costs for activation storage, the substantial computational overhead of the backward pass, the backward locking effect that inhibits parallelism, and the potential for gradient instability \cite{bengio2015towards, lillicrap2020backpropagation, jaderberg2017decoupled, hochreiter2001gradient}. Second, there is a distinct lack of systematic studies that rigorously compare emerging BP-free algorithms such as FF, CaFo, and MF against BP under controlled conditions, particularly those that evaluate both performance and direct energy efficiency metrics \cite{johnson2023inconsistency, strubell2019energy}.

This gap is exacerbated by a third issue: Architectural variations frequently confound existing comparisons. Studies often assess alternative algorithms on architectures that are distinct from standard BP configurations, making it difficult to disentangle the influence of the training algorithm from that of the architectural design \cite{desislavov2024energy, yarally2023uncovering}. A critical deficiency is the absence of comparisons where alternative algorithms are evaluated on their specific "native" architectures, and fair backpropagation baselines are constructed on those same architectures, a condition essential for an accurate assessment. Fourth, energy efficiency analysis is often inconsistent, relying on indirect proxies such as floating-point operations (FLOPs) rather than direct hardware monitoring. This practice can yield incomplete or misleading conclusions about the true energy profiles of different training methods \cite{rajput2023enhancing, charpentier2023accuracy, chung2024zeus}, and standardized reporting of estimated CO2e using tools such as CodeCarbon is often omitted. Finally, as a consequence of these issues, the specific trade-offs between classification performance and a comprehensive suite of efficiency metrics (energy, time, memory, GFLOPs, estimated CO2e) for these newer algorithms relative to established baselines of BP remain inadequately characterized, thus limiting their informed adoption \cite{yarally2023uncovering, narayanagowda2023watt}.

Addressing these deficiencies is imperative for the advancement of sustainable deep learning. Therefore, a standardized methodology is required to rigorously evaluate the performance and efficiency of emerging training algorithms against established baselines on equitable architectural grounds.

\subsection{Research Objectives and Questions}
\label{sec:objectives}

To address this need, the primary objective of this paper is to conduct a comprehensive and fair comparison of the Forward-Forward (FF), Cascaded-Forward (CaFo), and Mono-Forward (MF) algorithms against a standard backpropagation-based approach. The experimental design focuses on implementing each algorithm in its native network architecture and establishing a corresponding fair BP baseline that uses an identical architecture, thereby isolating the impact of the training method itself.

This paper seeks to provide empirical answers to the following research questions:
\begin{enumerate}
    \item How do the alternative training algorithms (FF, CaFo, and MF) compare to backpropagation in terms of classification accuracy and convergence rate on the MNIST, Fashion-MNIST, CIFAR-10, and CIFAR-100 datasets? This investigation is structured to evaluate FF foundationally on Fashion-MNIST, CaFo in its native CNN architecture, and MF in its native MLP architecture across all relevant datasets.

    \item What are the directly measured energy consumption profiles (via the NVIDIA Management Library (NVML)) of FF, CaFo, and MF compared to their fair backpropagation baselines during training on identical native architectures?

    \item How do key computational efficiency metrics differ between the alternative algorithms and their fair BP baselines under controlled architectural settings, including MLP structures native to MF?
    
    This analysis covers:
        \begin{itemize}
            \item Wall-clock training time.
            \item Peak GPU memory usage.
            \item Estimated forward-pass Giga Floating-Point Operations (GFLOPs) as a proxy for inference complexity.
            \item An estimation of the computational cost per update cycle for BP, illustrating the theoretical savings from eliminating the backward pass.
            \item Estimated Carbon Dioxide Equivalent (CO2e) calculated using the CodeCarbon library.
        \end{itemize}

    \item What specific energy performance trade-offs (including estimated CO2e) exist for CaFo and MF relative to backpropagation? How does FF's trade-off compare within its more limited experimental scope?
    
    \item How does the scalability, in terms of both performance and efficiency, of CaFo (on CNNs) and MF (on MLPs) vary across datasets of increasing complexity when compared to BP, and what are the implications for their potential deployment in energy-constrained environments?
\end{enumerate}

\subsection{Paper Contributions}
\label{sec:contributions}

This paper makes several distinct contributions to the field of energy-efficient deep learning:
\begin{itemize}
    \item \textbf{Rigorous Algorithm Replication:} The author faithfully reproduced the native architectures for the FF (MLP), CaFo (CNN, including both Rand CE and the implemented DFA CE variants), and MF (MLP) algorithms from their original descriptions.

    \item \textbf{Fair Comparative Framework:} By implementing both alternative algorithms and their backpropagation counterparts on identical network architectures and applying consistent practices such as early stopping based on validation performance, this paper establishes a framework that isolates the impact of the training algorithm itself.

    \item \textbf{Empirical Energy Efficiency Analysis:} Through direct hardware measurements using the NVIDIA NVML API and profiling tools, this work provides a detailed, multifaceted efficiency profile for each training method under fair conditions. This profile includes direct energy, time, peak memory, estimated forward-pass GFLOPs, and an estimated CO2e, alongside an estimation of BP's per-update computational cost to highlight algorithmic differences.

    \item \textbf{Trade-off Characterization:} This paper presents a thorough analysis of the trade-offs between classification performance, determined from the best model state identified by validation, and comprehensive energy efficiency, offering valuable insights for the practical deployment of deep learning models in energy-conscious settings.

    \item \textbf{Scalability Assessment:} Through systematic experiments on datasets of increasing complexity (MNIST, Fashion-MNIST, CIFAR-10, and CIFAR-100), the scalability and practical applicability of these novel training methods are evaluated.

    \item \textbf{Open Research Infrastructure:} To promote reproducibility and facilitate future research, all source code, structured configuration files (employing an inheritance-based YAML system), experimental protocols and implementations of measurement tools are publicly available in a GitHub repository, with additional details provided in the Appendix~\ref{app:code_repository}.
\end{itemize}

\subsection{Overview of Paper Structure}
\label{sec:structure}

This paper is organized into six chapters, each designed to build on the preceding one to construct a comprehensive investigation of energy-efficient training methods for DNNs:
\begin{itemize}
    \item \textbf{Chapter 1: Introduction} outlines the motivation, defines the problem statement, and details the research objectives, contributions, and general structure of the paper.

    \item \textbf{Chapter 2: Background and Literature Review} provides an overview of deep neural networks and the backpropagation algorithm, reviews pressing challenges in energy efficiency, and details the alternative training algorithms (FF, CaFo, MF) alongside the rationale for fair benchmarking. This chapter frames the selected algorithms within an evolutionary context, beginning with FF's foundational concepts, progressing to CaFo's structured block-wise methodology, and culminating in MF's recent advancements in both efficiency and performance.

    \item \textbf{Chapter 3: Methodology} describes the experimental design, dataset preprocessing, implementation details for the alternative algorithms and their corresponding fair backpropagation baselines, evaluation metrics, and hardware and software infrastructure. A key focus of this chapter is the establishment of a methodologically rigorous framework for ensuring fair comparisons.

    \item \textbf{Chapter 4: Experiments and Results} presents a detailed account of the empirical findings, including a full suite of performance and efficiency metrics for both the alternative algorithms and their fair BP baselines, supported by comparative analyses and statistical evaluations.

    \item \textbf{Chapter 5: Discussion} interprets the experimental results, examines the specific trade-offs between energy efficiency and model performance, and discusses the advantages, limitations, and practical implications of the investigated training methods.

    \item \textbf{Chapter 6: Conclusion and Future Work} summarizes the key findings, revisits the contributions of the paper, acknowledges its limitations, and proposes promising directions for future inquiry.
\end{itemize}
This structure is intended to guide the reader systematically through the context, methods, results, and implications of this research, which explores alternatives beyond conventional backpropagation to foster a more energy-efficient future for deep learning.
\section{Literature Review}
\label{chap:background}

This chapter establishes the theoretical foundations and contextual framework essential for interpreting the comparative analysis presented in this paper. It commences with an exposition of deep neural networks and the canonical backpropagation algorithm, before proceeding to a discussion of the energy efficiency challenges endemic to deep learning. The chapter subsequently provides a detailed examination of three alternative training paradigms: the Forward-Forward (FF), Cascaded-Forward (CaFo), and Mono-Forward (MF) algorithms, which constitute the central focus of this research. These algorithms are situated within an evolutionary narrative of BP-free learning, where each method either builds upon or diverges from preceding concepts to resolve specific limitations. The chapter culminates by exploring established benchmarking practices for ensuring fair comparison against backpropagation and identifying the specific gaps in the existing literature that this paper seeks to address, with a pronounced emphasis on methodological rigor.

\subsection{Deep Neural Networks and Backpropagation}
\label{sec:dnn_bp}

Deep neural networks (DNNs) have become established as a class of powerful models capable of discerning complex patterns from large-scale datasets, thereby facilitating breakthrough performance across a multitude of domains \cite{goodfellow2016deep}. Fundamentally, DNNs are constructed from multiple layers of interconnected computational units, or neurons, which transform input data through a sequence of non-linear mappings to produce semantically meaningful outputs. The efficacy of these networks is largely attributable to their hierarchical architecture, which enables the learning of progressively more abstract data representations.

The canonical training methodology for DNNs is the backpropagation (BP) algorithm \cite{rumelhart1986learning}, a technique that has retained its dominance for over three decades. Backpropagation operates via two distinct phases: a forward pass, during which input signals traverse the network to generate predictions, and a backward pass, where error gradients are propagated from the output layer to preceding layers to guide the adjustment of weights. This process is formally defined as follows.

For a neural network comprising $L$ layers, let $\mathbf{x}_l$ denote the activations in layer $l$, $\mathbf{W}_l$ the weight matrix connecting layer $l-1$ to layer $l$, $\mathbf{b}_l$ the bias vector in layer $l$ and $f_l$ the corresponding activation function. The forward pass computes:

\begin{equation}
\mathbf{z}_l = \mathbf{W}_l\mathbf{x}_{l-1} + \mathbf{b}_l
\tag{2.1}
\label{eq:bp_forward_z_chap2}
\end{equation}

\begin{equation}
\mathbf{x}_l = f_l(\mathbf{z}_l)
\tag{2.2}
\label{eq:bp_forward_x_chap2}
\end{equation}

In the subsequent backward pass, for a given loss function $\mathcal{L}$ that measures the discrepancy between the predictions of the network $\mathbf{x}_L$ and the target outputs $\mathbf{y}$, the gradients are computed. The process begins at the output layer:

\begin{equation}
\frac{\partial \mathcal{L}}{\partial \mathbf{z}_L} = \frac{\partial \mathcal{L}}{\partial \mathbf{x}_L} \odot f'_L(\mathbf{z}_L)
\tag{2.3}
\label{eq:bp_backward_L_chap2}
\end{equation}

For any hidden layer $l$ (where $1 \le l < L$), the gradient is propagated backward:

\begin{equation}
\frac{\partial \mathcal{L}}{\partial \mathbf{z}_l} = \left(\mathbf{W}_{l+1}^T \frac{\partial \mathcal{L}}{\partial \mathbf{z}_{l+1}}\right) \odot f'_l(\mathbf{z}_l)
\tag{2.4}
\label{eq:bp_backward_l_chap2}
\end{equation}

Finally, the gradient with respect to the weights of the layer $l$ is calculated as:

\begin{equation}
\frac{\partial \mathcal{L}}{\partial \mathbf{W}_l} = \frac{\partial \mathcal{L}}{\partial \mathbf{z}_l} \mathbf{x}_{l-1}^T.
\tag{2.5}
\label{eq:bp_backward_W_chap2}
\end{equation}

As detailed in Equations (\ref{eq:bp_forward_z_chap2}) and (\ref{eq:bp_forward_x_chap2}), the forward pass entails the computation of pre-activation values followed by the application of activation functions. The backward pass, initiated by Equation (\ref{eq:bp_backward_L_chap2}) in the output layer, propagates the gradient through each hidden layer via the chain rule, as shown in Equation (\ref{eq:bp_backward_l_chap2}). Equation (\ref{eq:bp_backward_W_chap2}) then specifies how the gradients with respect to the weights are determined.

Despite its demonstrated efficacy, backpropagation is beset by several inherent limitations. First, it requires the storage of all intermediate activations from the forward pass for subsequent use in the backward pass, a requirement that leads to substantial memory consumption scaling linearly with network depth \cite{chen2016training}. This consequently increases the energy expenditure associated with data storage and movement. Second, the algorithm's sequential nature gives rise to a computational bottleneck known as backward locking, wherein weight updates for a given layer are deferred until the completion of both the forward and backward passes, including gradient computations from all subsequent layers. This dependency obstructs layer-parallel updates and reduces opportunities for holistic training parallelization \cite{jaderberg2017decoupled}. Third, the algorithm is susceptible to the \emph{vanishing or exploding gradients} problem, where backpropagated error signals either diminish exponentially, rendering updates to early layers ineffective, or grow uncontrollably, destabilizing the training process. This phenomenon is primarily caused by the repeated multiplication of Jacobian factors (related to $f'_l$ and $\mathbf{W}_{l+1}^T$ in Equation (\ref{eq:bp_backward_l_chap2})) through the deep network chain \cite{hochreiter2001gradient}.

In addition, backpropagation has faced criticism for its \textbf{lack of biological plausibility}. The algorithm relies on precise, symmetric weight transport between forward and backward passes (i.e., the use of $\mathbf{W}_{l+1}^T$ in Eq. \ref{eq:bp_backward_l_chap2}) is a requirement for which there is no substantial evidence in biological neural circuits \cite{lillicrap2020backpropagation}. The propagation of global error signals also stands in contrast to the predominantly local learning mechanisms observed in biological neurons \cite{bengio2015towards}.

These constraints have catalyzed research into alternative training methodologies that seek to ameliorate the deficiencies of backpropagation while preserving competitive levels of performance. Of particular relevance are approaches that obviate the backward pass entirely, offering potential enhancements in memory efficiency, training velocity, and energy consumption.

\subsection{Energy Efficiency in Deep Learning}
\label{sec:energy_efficiency_chap2}

The exponential scaling of computational demands for training contemporary deep neural networks has raised significant concerns about energy consumption and environmental impact. Seminal studies have shown that the training of a single large language model can generate carbon emissions comparable to the lifetime emissions of five automobiles \cite{strubell2019energy}. This phenomenon is symptomatic of a wider increase in the energy requirements of information and communication technologies (ICT), with reports indicating that data centers and communication networks are now responsible for 2 to 3\% of global electricity consumption and approximately 1\% of greenhouse gas emissions \cite{wef2024growing}. Although initial projections from 2015 suggested that ICT could constitute a significant portion of global electricity usage by 2030 \cite{andrae2015global}, more recent analyses from the Oxford Institute for Energy Studies forecast a base case scenario where data centers alone will consume 1,135 TWh in the same year \cite{oxford2024global}. The combined electricity demand from data centers, cryptocurrencies and AI is expected to increase substantially by 2026 \cite{wef2024growing}, highlighting the need for energy-efficient deep learning paradigms.

The energy consumption profile of deep learning is multifaceted. A primary contributor is the computational workload of the training process itself, particularly matrix multiplications and evaluations of activation functions, which require considerable computational power. Floating-point operations (FLOPs) are commonly employed as an abstract metric to estimate these computational costs.

However, it is crucial to recognize that, while FLOPs provide a useful measure of computational complexity, they do not directly quantify energy consumption. FLOPs represent a simple tally of arithmetic operations, whereas the actual energy expenditure per operation can exhibit significant variance contingent on hardware specifics such as processor architecture, numerical precision, and power state. Furthermore, a substantial fraction of energy, often rivaling or exceeding that of computation, is consumed by data movement across the memory hierarchy (e.g., DRAM, cache, registers) and interconnects, a factor that correlates poorly with FLOP counts \cite{horowitz2014computing}.

Moreover, the notion of a fixed FLOP count for a fundamental operation is itself being challenged by AI-driven algorithmic discovery. For example, DeepMind AlphaTensor, a deep reinforcement learning agent, discovered novel matrix multiplication algorithms that are asymptotically more efficient than established methods like Strassen's algorithm for certain matrix sizes \cite{fawzi2022discovering}. By identifying a method to multiply 4x4 matrices in 47 steps instead of Strassen's 49, it achieved the first such improvement in more than fifty years, demonstrating that the foundational number of operations for a task is not an immutable baseline. This development serves as a powerful parallel to the objective of this paper: just as AI can optimize a core \emph{operation} such as matrix multiplication, this research investigates the replacement of the core training \emph{algorithm}, backpropagation, to achieve greater efficiency. Ultimately, the dynamic nature of algorithmic efficiency underscores the limitations of relying solely on theoretical proxies and reinforces the methodological decision in this work to rely on direct, empirical measurements of energy, time, and memory for a robust and faithful comparison.

This need for a nuanced and empirical approach is especially clear when evaluating backpropagation (BP). An analysis restricted to forward-pass FLOPs is insufficient, as a complete BP update cycle encompasses a forward-pass, a backward-pass for gradient computation, and subsequent weight updates. As a widely accepted heuristic, the backward pass typically entails approximately twice the computational workload (FLOPs) of the forward pass for standard network layers \cite{kaplan2020scaling}. Consequently, the total theoretical computational cost of a single BP update cycle, which this paper terms $F_{BP\_update}$, can be estimated relative to the forward pass cost ($F_{fwd}$) as follows:
\begin{equation}
F_{BP\_update} \approx F_{fwd} (\text{forward}) + 2 \times F_{fwd} (\text{backward}) = 3 \times F_{fwd}
\label{eq:bp_update_cost}
\end{equation}
BP-free methods, by their design, eliminate the $2 \times F_{fwd}$ cost of the backward pass. Although this 3x estimate serves as a useful theoretical benchmark, its ultimate impact on overall training efficiency is modulated by factors such as convergence speed and hardware utilization, reinforcing the need for direct time and energy measurements. In this paper, both the estimated forward pass GFLOPs ($F_{fwd}$) for all algorithms and the estimated BP update cycle GFLOPs ($F_{BP\_update}$) are reported to illustrate this computational difference.

More direct and accurate measurements of energy consumption can be achieved through specialized APIs and instrumentation. The NVIDIA Management Library (NVML) API, for instance, provides real-time power readings from GPU devices, facilitating precise energy usage estimates during training. Tools such as CarbonTracker \cite{anthony2020carbontracker} and CodeCarbon \cite{codecarbon2021} utilize these measurements, integrating factors such as runtime and regional carbon intensity of the electricity grid to estimate the associated carbon footprint, typically reported in Carbon Dioxide Equivalent (CO2e) \cite{codecarbon2021}.

Despite the availability of these tools, systematic reporting of energy metrics remains inconsistent within the deep learning research community \cite{henderson2020towards}. Many studies prioritize model accuracy and convergence speed, relegating energy considerations to a secondary status. When energy metrics are reported, they frequently lack methodological standardization, which complicates direct comparisons between different approaches.

The concept of "Green AI" \cite{schwartz2020green} has arisen in response to these challenges, advocating for the elevation of efficiency to a primary evaluation criterion along with precision. This paradigm encourages the development of models that achieve state-of-the-art performance while minimizing resource consumption, thereby moving away from a paradigm of pursuing marginal accuracy gains with ever larger models at a significant environmental cost.

It is also pertinent to note that many existing techniques for mitigating the energy cost of deep learning, such as network pruning, quantization \cite{jacob2018quantization}, knowledge distillation, and the development of inherently efficient architectures like MobileNets \cite{howard2017mobilenets} and EfficientNets \cite{tan2019efficientnet}, primarily target the model's structure or its inference phase. Although these methods are valuable, they are largely \emph{orthogonal} to the choice of the core training algorithm that dictates how weights are updated. This paper is specifically focused on the potential energy savings that can be achieved by fundamentally altering this training algorithm and moving beyond backpropagation.

Alternative training algorithms that eliminate or modify the backpropagation process represent a promising avenue to improve energy efficiency during the \emph{training} phase. By potentially reducing memory overhead, enabling superior parallelization, and minimizing the data movement and computation associated with the backward pass, these approaches have the potential to substantially decrease the energy footprint of deep learning training while maintaining competitive performance.

\subsection{Alternative Training Algorithms: An Evolutionary Perspective}
\label{sec:alternative_algorithms_chap2}

This section investigates three innovative training algorithms: Forward-Forward (FF), Cascaded-Forward (CaFo), and Mono-Forward (MF), which are designed to transcend the limitations of backpropagation. These methods are presented as stages in an evolutionary progression towards BP-free learning. FF established the fundamental concept of local, forward-only training. CaFo advanced this idea by introducing a structured block-wise framework to integrate supervisory signals more directly, although with its own set of complexities. MF represents a more recent refinement, developed to achieve high performance and efficiency through a distinct local learning mechanism. A common characteristic shared by all three algorithms is the elimination of the global backward pass; they rely instead on local learning rules that update weights based solely on information available within each layer.

\subsubsection{Forward-Forward (FF) Algorithm}
\label{subsec:ff_algorithm_chap2}

Proposed by Geoffrey Hinton in 2022, the FF algorithm \cite{hinton2022forward} constitutes a radical departure from conventional backpropagation, motivated by considerations of biological plausibility and the potential for greater computational efficiency. Its central innovation was to demonstrate that effective learning can be achieved through purely local objectives and two distinct forward passes: one for "positive" (real) data and another for "negative" (typically artificially generated) data, thereby obviating the need to propagate error gradients backwards. Each layer is trained locally to differentiate between these data types by optimizing a layer-specific "goodness" metric \cite{hinton2022forward}.

The core mechanism of the algorithm is as follows:
\begin{enumerate}
    \item \textbf{Data Presentation:} In a supervised context, positive data is formulated by presenting an input, such as an image, paired with its correct label. This pairing is often accomplished by embedding the label directly into the input representation, for example, by replacing the first few pixels. Subsequently, negative data is created by pairing the same input with an incorrect label \cite{hinton2022forward}.
    \item \textbf{Goodness Function:} Each layer $l$ calculates a goodness score derived from its activation pattern $\mathbf{x}_l$. A common formulation for this score is the sum of squared activations: $G_l(\mathbf{x}_l) = \sum_i x_{l,i}^2$. The weights of the layer are then adjusted to increase this goodness for positive samples while decreasing it for negative samples \cite{hinton2022forward}.
    \item \textbf{Local Objective:} The learning objective for a layer with parameters $\boldsymbol{\theta}_l$ is designed to maximize the difference in goodness between positive ($\mathbf{x}_l^+$) and negative ($\mathbf{x}_l^-$) inputs. 

This is frequently framed using a logistic loss function applied to this difference, which may be shifted by a threshold $\theta$:
        \begin{equation}
        \mathcal{L}_l(\boldsymbol{\theta}_l) = \log(1 + e^{-(G_l(\mathbf{x}_l^+) - \theta)}) + \log(1 + e^{(G_l(\mathbf{x}_l^-) - \theta)})
        \tag{2.6} \label{eq:ff_loss_chap2}
        \end{equation}
        where the optimization goal is to minimize $\mathcal{L}_l$.
    \item \textbf{Layer Normalization:} Before propagating activations $\mathbf{x}_l$ to the subsequent layer $l+1$, FF applies layer normalization \cite{ba2016layer}, which is often simplified to length normalization by dividing by the L2 norm. This step is of critical importance: it removes the magnitude information that defined the goodness in layer $l$, thereby compelling subsequent layers to learn new discriminative features based on the orientation or relative activities of the activation vector, rather than merely inheriting the goodness signal \cite{hinton2022forward}.
\end{enumerate}

This forward-only, local learning paradigm presents several potential advantages pertinent to energy-efficient training:
\begin{itemize}
    \item \textbf{Reduced Memory Footprint:} By eliminating the backward pass, FF obviates the need to store intermediate activations for gradient computation. Memory requirements scale with the size of the layer rather than the depth of the network, which offers a significant potential for savings relative to BP, particularly in deep networks \cite{hinton2022forward}.
    \item \textbf{Enhanced Parallelizability:} As each layer updates its weights locally and independently without backward locking, FF has the potential to facilitate greater parallelism across layers or even enable pipelined processing of data streams \cite{hinton2022forward}.
    \item \textbf{Avoidance of Gradient Issues:} The local nature of the objective function inherently bypasses the issues of vanishing or exploding gradients associated with deep backpropagation chains \cite{hinton2022forward}.
\end{itemize}

However, FF also presents a unique set of characteristics and challenges:
\begin{itemize}
    \item \textbf{Native Architecture (MLP Focus):} It is essential to note that Hinton's original research mainly focused on developing and evaluating FF on MLPs with fully-connected layers, such as four-layer networks with 2000 ReLU neurons each for MNIST and Fashion-MNIST \cite{hinton2022forward}. Hinton himself noted difficulties in adapting FF to CNNs. This paper adheres to the native MLP architecture for FF evaluation, specifically a 4$\times$2000 network for Fashion-MNIST as per the research plan.
    \item \textbf{Inference Procedure:} The standard inference process in FF requires multiple forward passes. To classify a given input, the network computes the total goodness, summed across relevant layers, for the input paired with each possible class label. The label that produces the highest total goodness is then selected as the prediction \cite{hinton2022forward}.
    \item \textbf{Performance Relative to Backpropagation:} The results of Hinton's work and subsequent studies generally indicate that FF, in its native MLP form, achieves reasonable accuracy on datasets like MNIST but tends to lag behind standard BP, particularly on more complex datasets such as CIFAR-10 \cite{hinton2022forward}. Convergence is also typically observed to be slower \cite{hinton2022forward}.
    \item \textbf{Hyperparameter Sensitivity:} The algorithm can exhibit sensitivity to hyperparameters such as learning rates and goodness thresholds ($\theta$ in Eq. \ref{eq:ff_loss_chap2}), and the specifics of the normalization procedure \cite{hinton2022forward}.
\end{itemize}

From a biological point of view, FF is considered as more plausible than BP because it employs local learning rules, avoids the need for precise symmetric weight transport, and does not require long-term storage of activations \cite{hinton2022forward}.

\textbf{In the context of this paper}, FF serves as a foundational example of a BP-free algorithm, representing a critical conceptual advance. Its potential for reduced memory usage and enhanced parallelizability makes it highly relevant to the study of energy efficiency, although its practical efficiency in the tests conducted for this paper proved to be limited. By evaluating FF on its native MLP architecture against an identical MLP trained with BP, this research aims to isolate the impact of the FF mechanism on both performance and energy metrics.

\subsubsection{Cascaded-Forward (CaFo) Algorithm}
\label{subsec:cafo_algorithm_chap2}

The CaFo algorithm \cite{zhao2023cafo}, proposed by Zhao et al. in 2023, builds upon concepts introduced in FF and aims to address some of the limitations of Hinton's approach, including its reliance on negative samples and the indirect nature of the "goodness" signal for supervision. CaFo introduces a structured, block-wise training process designed to improve stability, convergence, and accuracy while retaining the advantages of BP-free learning.

CaFo is characterized by several key features:
\begin{enumerate}
    \item \textbf{Cascaded Neural Blocks:} The architecture comprises sequential "neural blocks," each containing standard components such as convolutional or fully-connected layers, normalization, and activation functions that progressively extract features \cite{zhao2023cafo}.
    \item \textbf{Layer-wise Predictors (Auxiliary Classifiers):} A dedicated predictor, typically a shallow fully-connected layer followed by a softmax function, is attached to each neural block. This predictor maps the block's output features directly to task-specific outputs, such as class probabilities, thereby providing a more direct supervisory signal at multiple depths within the network \cite{zhao2023cafo}.
    \item \textbf{Independent Component Training:} The predictors can be trained independently, often using the outputs of pre-trained or even randomly initialized neural blocks. The blocks themselves can be trained using non-BP methods such as direct feedback alignment (DFA) \cite{nokland2016direct} or remain fixed \cite{zhao2023cafo}.
    \item \textbf{Single Input Stream:} In contrast to the positive/negative passes of FF, CaFo utilizes a single input stream and employs ground truth labels for local predictor losses, which simplifies data handling \cite{zhao2023cafo}.
    \item \textbf{Knowledge Integration during Inference:} The final prediction is typically derived by aggregating the outputs from all predictors, for example, by averaging their probability distributions. This approach utilizes information from multiple levels of representation in a single forward pass through the architecture \cite{zhao2023cafo}.
\end{enumerate}

\textbf{Native Architecture and Implementation for Fair Benchmarking:}
As specified in the CaFo publication \cite{zhao2023cafo}, and pertinent to the methodology of this paper, the authors proposed and evaluated a specific CNN variant for image classification tasks. This native architecture, which was used for datasets including Fashion-MNIST and CIFAR-10/100, involves several cascaded convolutional blocks. For the experiments in this paper, this research adheres strictly to this native CNN architecture. It consists of \textbf{three blocks}, each containing a \textbf{Conv2D(3x3), ReLU, MaxPool(2x2), and BatchNorm} layer, with progressively increasing channel depths (e.g., \textbf{32, 128, 512 channels}). The output feature map of each block is flattened before being fed into its dedicated predictor (e.g., an \textbf{FC layer + Softmax}). The corresponding fair BP baseline replicates this exact CNN architecture and is trained end-to-end.

\pagebreak 

\textbf{Training Methodology and Efficiency Aspects:}
CaFo training involves distinct phases, which have been implemented as part of this work:
\begin{itemize}
    \item \textbf{Neural Block Training (Optional: CaFo-DFA Variant):} The neural blocks can be randomly initialized and kept fixed, which constitutes the \textit{CaFo-Rand} variant. Alternatively, they can be trained using a BP-free method such as direct feedback alignment (DFA) \cite{nokland2016direct}. DFA offers a solution to the biological implausibility of backpropagation by eliminating the need for symmetric weight transport. Instead of propagating the error backward using the transpose of the forward weight matrices, DFA generates a synthetic gradient for each layer. This is achieved by projecting the global error signal from the output layer directly back to each hidden layer via a fixed, random feedback matrix. As this matrix is not learned and is independent of the forward weights, it allows each layer to compute an approximate weight update locally, thus avoiding the backward locking problem. Although often yielding lower accuracy than BP, its mechanism serves as an important benchmark for other BP-free algorithms \cite[Table 3]{gong2025mono}. This DFA block training mechanism has been implemented, which uses an approximation of DFA for convolutional layers, as a configurable option for evaluating the \textit{CaFo-DFA} variant.
    
    \item \textbf{Predictor Training:} Each predictor is trained independently on the output of its corresponding block. CaFo supports multiple loss functions, including Mean Squared Error (MSE), Cross-Entropy (CE), and Sparsemax Loss (SL) \cite{zhao2023cafo}. It should be noted that using the MSE loss for predictors permits a \textbf{closed-form solution} via linear regression, allowing exceptionally fast predictor training once the block outputs ($\mathbf{H}$) are computed \cite{zhao2023cafo}. Although the MSE loss variant offers this exceptional training speed, this work focuses on the Cross Entropy (CE) variants, as they are reported to yield higher classification accuracy, providing a more stringent test against backpropagation's performance \cite{zhao2023cafo}. Training with CE or SL involves gradient descent but is based solely on local gradients computed through the shallow predictor.
\end{itemize}
This modular training approach, which eliminates the deep backward pass, suggests potential advantages in terms of reduced memory usage and faster training times, particularly when using MSE loss, making it highly relevant for energy efficiency analysis. The use of DFA for block training offers an additional BP-free pathway within the CaFo framework.

\textbf{Performance and Trade-offs:}
The results presented in \cite{zhao2023cafo} indicate that CaFo significantly outperforms FF on benchmarks such as MNIST and CIFAR-10/100. Although it still generally underperforms an end-to-end BP-trained model on the same architectures, the performance gap reported for CaFo, especially with CE or SL loss and DFA-trained blocks, is smaller than that of FF \cite{zhao2023cafo}. The choice of loss function (MSE vs. CE/SL) presents a trade-off between speed and accuracy \cite{zhao2023cafo}. The experiments in this paper are designed to quantify these performance and efficiency trade-offs against a fair BP baseline, with a primary focus on the Rand-CE variant, while also acknowledging the implemented DFA capability.

\textbf{Biological Plausibility:}
Similar to FF, CaFo is argued to be more biologically plausible than BP due to its avoidance of weight transport, achieved through the use of DFA or local predictors, and its lack of global error propagation \cite{zhao2023cafo}.

\textbf{In summary}, CaFo represents an evolutionary step from FF, offering a structured forward-only learning approach with the potential for improved accuracy and specific efficiency advantages. Its distinct native architectures and training options, including the Rand-CE and DFA-CE variants implemented, establish it as a key algorithm for the comparative study in this paper.

\pagebreak

\subsubsection{Mono-Forward (MF) Algorithm}
\label{subsec:mf_algorithm_chap2}

The MF algorithm, introduced by Gong, Li, and Abdulla \cite{gong2025mono}, is a recent BP-free training method that builds on the core principle of local, forward-only learning. It employs a novel mechanism, local projection matrices, with the objective of achieving performance that can match or even surpass that of backpropagation, particularly on MLP architectures, while offering substantial benefits in memory efficiency and parallelizability. MF simplifies certain aspects of both FF, by eliminating the need for negative samples, and CaFo, by replacing complex cascaded predictor training with a direct local loss computation on each layer's activations.

\textbf{Core Mechanism: Local Learning with Projection Matrices}
The key innovation of MF lies in its layer-wise optimization process, which utilizes a dedicated \emph{projection matrix} $\mathbf{M}_i$ in each hidden layer $i$. This matrix, with dimensions $m \times n$ (where $m$ is the number of classes and $n$ is the number of neurons in the layer), maps the layer's activation vector $\mathbf{a}_i$ (of size $1 \times n$ for a single sample, or $B \times n$ for a batch of size $B$) directly to category-specific "goodness" scores $\mathbf{G}_i$ (of size $1 \times m$ or $B \times m$):

\begin{equation}
\mathbf{G}_i = \mathbf{a}_i \mathbf{M}_i^T
\tag{2.7} \label{eq:mf_goodness_chap2}
\end{equation}

MF then applies a standard cross-entropy loss, computed locally at each layer $i$, between the softmax of these goodness scores and the true labels $\mathbf{y}$:

\begin{equation}
\mathcal{L}_i = \text{CrossEntropy}(\text{softmax}(\mathbf{G}_i), \mathbf{y})
\tag{2.8} \label{eq:mf_loss_chap2}
\end{equation}
Critically, both the layer's weights $\mathbf{W}_i$ and its projection matrix $\mathbf{M}_i$ are updated using gradients derived exclusively from this local loss $\mathcal{L}_i$:

\begin{align}
\mathbf{W}_i &\leftarrow \mathbf{W}_i - \eta \frac{\partial \mathcal{L}_i}{\partial \mathbf{W}_i} \tag{2.9} \label{eq:mf_update_W_chap2} \\
\mathbf{M}_i &\leftarrow \mathbf{M}_i - \eta \frac{\partial \mathcal{L}_i}{\partial \mathbf{M}_i} \tag{2.10} \label{eq:mf_update_M_chap2}
\end{align}
where $\eta$ represents the learning rate. The gradient for the projection matrix, $\frac{\partial \mathcal{L}_i}{\partial \mathbf{M}_i}$, is computed directly. The gradient for the layer weights is determined via a local "mini-backpropagation" through the layer itself, as defined by the chain rule:

\begin{equation}
\frac{\partial \mathcal{L}_i}{\partial \mathbf{W}_i} = \frac{\partial \mathcal{L}_i}{\partial \mathbf{G}_i} \frac{\partial \mathbf{G}_i}{\partial \mathbf{a}_i} \frac{\partial \mathbf{a}_i}{\partial \mathbf{z}_i} \frac{\partial \mathbf{z}_i}{\partial \mathbf{W}_i}
\tag{2.11} \label{eq:mf_local_chain_rule}
\end{equation}
where $\mathbf{z}_i$ is the pre-activation value. This local mechanism effectively avoids backward locking and the problems associated with global gradient propagation \cite{gong2025mono}.

\textbf{Native Architectures and Fair Benchmarking Relevance}
The original MF paper \cite{gong2025mono} primarily evaluated the algorithm using MLPs on standard image classification benchmarks. The native architectures specified in the paper, which are relevant to this paper, are:
\begin{itemize}
    \item \textbf{MNIST \& Fashion-MNIST:} A 2-hidden-layer MLP with 1000 ReLU neurons per hidden layer (a \textbf{2$\times$1000 ReLU MLP}).
    \item \textbf{CIFAR-10 and CIFAR-100:} A 3-hidden-layer MLP with 2000 ReLU neurons per hidden layer (a \textbf{3$\times$2000 ReLU MLP}).
\end{itemize}
Each hidden layer in these architectures incorporates its own projection matrix $\mathbf{M}_i$, which is initialized using a method such as Kaiming uniform initialization \cite{he2015delving}, and is trained using the local loss $\mathcal{L}_i$. In the methodology of this paper, these specific MLPs will be implemented for MF, and the fair BP baselines will utilize \emph{identical} MLP structures trained end-to-end.

\pagebreak
\textbf{Prediction Methods}
MF supports two distinct approaches for inference \cite{gong2025mono}:
\begin{enumerate}
    \item \textbf{FF-style Aggregation:} The goodness scores ($\mathbf{G}_i$) from all layers are aggregated, for instance, by summation, and the class corresponding to the highest aggregate score is predicted.
    \item \textbf{BP-style Final Layer:} Only the goodness scores from the final hidden layer ($\mathbf{G}_L$) are used for prediction via a softmax function. This method is computationally analogous to standard BP inference, requiring a single forward pass.
\end{enumerate}
The original paper suggests that BP-style inference is both efficient and performs well \cite{gong2025mono}.

\textbf{Potential for Energy Efficiency}
The design of the MF algorithm holds significant promise for energy efficiency:
\begin{itemize}
    \item \textbf{Memory Usage:} The elimination of the backward pass obviates the need to store intermediate activations. The MF paper provides empirical evidence of substantially lower peak memory usage compared to BP \cite[Fig. 1-3]{gong2025mono}.
    \item \textbf{Training Time:} The layer-wise independence of MF removes the backward locking constraint, which enables the potential layer-parallelism. The paper reports that MF trains significantly faster per epoch than BP on comparable MLP architectures \cite[Table 5]{gong2025mono}.
    \item \textbf{FLOPs:} While MF eliminates the FLOPs associated with the backward pass, it introduces additional FLOPs for projection matrix computations. The net effect on computational load requires empirical measurement.
    \item \textbf{Energy Consumption:} The combination of lower memory requirements, potentially faster training, and parallelizable computations is expected to result in a reduced overall energy consumption. Direct measurements using NVML are necessary to verify this.
\end{itemize}
It should be noted that the projection matrices $\mathbf{M}_i$ add to the parameter count compared to a standard BP network, leading to a slight increase in model size \cite{gong2025mono}.

\textbf{Performance and Convergence}
Crucially, Gong et al. \cite{gong2025mono} report that MF consistently \textbf{matched or surpassed the accuracy of BP} on identical MLP architectures across the MNIST, Fashion-MNIST, CIFAR-10, and CIFAR-100 datasets, while also significantly outperforming FF, FA, and DFA \cite[Table 3]{gong2025mono}. MF also demonstrated remarkable convergence stability \cite[Fig. 5]{gong2025mono}.

\textbf{Biological Plausibility and Modularity}
MF aligns more closely with the principles of biological learning than BP, due to its use of local updates and the absence of weight transport \cite{gong2025mono}. The independence of its layers also confers a high degree of modularity \cite[Section 3.3]{gong2025mono}.

\textbf{In summary}, MF represents a significant recent advancement in the field of BP-free learning. It offers a compelling alternative that has the potential to deliver superior performance along with substantial efficiency advantages by directly mapping local layer activations to class-specific scores for local loss computation. The evaluation of MF on its native MLP architectures will be a primary focus of this paper's comparative analysis against fair BP baselines.

\subsection{Benchmarking Against Backpropagation and the Need for Fairness}
\label{sec:fair_benchmarking_chap2}
A rigorous evaluation of the performance and energy efficiency of alternative training algorithms relative to backpropagation requires a fair benchmarking methodology. However, a review of the literature reveals that comparisons are often confounded by several factors. The most prominent of these is the network architecture; discrepancies in layer types, depth, width, connectivity, and activation functions can exert a dramatic influence on both performance and energy consumption, independent of the training algorithm used \cite{desislavov2024energy, yarally2023uncovering}.

Previous studies comparing alternative algorithms often lacked the strict consistency required for a truly equitable assessment. For instance, comparisons frequently use standard BP architectures that may not replicate the potentially non-standard native architectures of the alternative algorithms, thereby hindering a fair assessment of the algorithmic impact. Furthermore, best practices such as systematic hyperparameter tuning and consistent validation-based early stopping are not always universally applied across all compared methods \cite{hinton2022forward}, making it difficult to disentangle algorithmic effects from those specific to a particular experimental setup. Research focused on energy efficiency has also highlighted the sensitivity of such measurements to architectural and configuration details \cite{garcia2019estimation}, further underscoring the necessity for the architecturally fair and methodologically rigorous comparison that this paper employs.
Addressing these issues requires a framework built upon several principles of fairness:
\begin{itemize}
\item \textbf{Architectural Parity:} The alternative algorithm and its backpropagation baseline should be evaluated on identical network structures.
\item \textbf{Systematic Optimization:} All the compared methods should undergo a rigorous, equivalent hyperparameter tuning to ensure that each performs at its potential.
\item \textbf{Consistent Evaluation Criteria:} Practices like early stopping based on validation performance must be universally applied to avoid penalizing algorithms with different convergence dynamics.
\item \textbf{Comprehensive and Consistent Metrics:} Efficiency should be evaluated using direct hardware measurements rather than relying solely on proxies.
\end{itemize}
The absence of a standardized methodology that incorporates these principles represents a significant gap in the literature. By adopting a fair benchmarking approach, which is centered on native architectures and consistent training practices, this paper aims to provide a robust assessment of the true performance and energy trade-offs offered by FF, CaFo, and MF relative to standard backpropagation.

\subsection{Summary of Related Work and Identification of Research Gaps}
\label{sec:research_gaps_chap2}
This chapter has reviewed the foundational principles of DNN training with backpropagation, the significant challenge of energy efficiency in deep learning, and three promising BP-free alternatives: the Forward-Forward (FF), Cascaded-Forward (CaFo), and Mono-Forward (MF) algorithms, framed within an evolutionary context.
Key conclusions drawn from the literature are as follows:
\begin{enumerate}
\item Backpropagation, despite its dominance, is characterized by limitations related to memory consumption, parallelism due to backward locking, gradient stability, and biological plausibility \cite{lillicrap2020backpropagation, bengio2015towards, jaderberg2017decoupled}.
\item The energy cost associated with training large-scale DNNs is substantial and continues to grow, requiring research into more efficient training methodologies \cite{strubell2019energy, patterson2021carbon, wef2024growing}.
\item Alternative algorithms such as FF \cite{hinton2022forward}, CaFo \cite{zhao2023cafo}, and MF \cite{gong2025mono} offer potential solutions by eliminating the backward pass and utilizing local learning rules.
\item These alternatives are often developed and evaluated on specific "native" architectures, such as MLPs for FF and MF or block-wise CNNs for CaFo, a factor critical for fair assessment \cite{hinton2022forward, zhao2023cafo, gong2025mono}.
\end{enumerate}
\pagebreak

Despite this progress, several research gaps persist, providing the motivation for this paper:
\begin{enumerate}
\item \textbf{Absence of a Fair and Holistic Comparative Framework:} There is a scarcity of studies that systematically compare emerging BP-free algorithms against backpropagation under truly equitable conditions. A rigorous framework would require implementing algorithms on their respective native architectures, establishing architecturally identical BP baselines, and applying consistent best practices such as universal hyperparameter optimization and validation-based early stopping to all methods.
\item \textbf{Inconsistent and Indirect Efficiency Measurement:} Many studies lack direct hardware-level energy measurements, often relying on proxies like FLOPs which may not reflect true energy profiles \cite{charpentier2023accuracy}. The systematic and transparent reporting of a comprehensive suite of metrics, including direct energy consumption (Wh), training time, peak memory, and estimated CO2e, is rarely performed for these newer algorithms against fairly-optimized baselines \cite{henderson2020towards}.
\item \textbf{Uncharacterized Performance-Efficiency Trade-offs:} As a consequence of the above, the specific trade-offs between classification accuracy and a full suite of measured efficiency metrics remain inadequately characterized for algorithms such as CaFo and MF when compared to rigorously-optimized BP baselines \cite{narayanagowda2023watt, yarally2023uncovering}.
\item \textbf{Unknown Scalability on Native Architectures:} It is not well understood how the performance and efficiency of algorithms like CaFo and MF, on their native architectures, scale across datasets of increasing complexity when compared to fair and optimized BP baselines.
\end{enumerate}
This paper addresses these gaps by conducting a rigorous comparative empirical study. It focuses on the performance and, critically, the energy efficiency of FF, CaFo, and MF against fair BP baselines implemented on their identical native architectures, where all methods benefit from systematic hyperparameter tuning and early stopping. The methodology detailed in Chapter \ref{chap:methodology} is designed to facilitate this fair comparison through direct hardware measurements and computational profiling.
\section{Methodology}
\label{chap:methodology}

This chapter details the research methodology used to investigate the research questions articulated in Chapter \ref{chap:introduction}. The author outlines the experimental design, the datasets utilized, the specific implementations of the alternative algorithms on their native architectures, and the corresponding configurations for fair BP baselines. Furthermore, this chapter describes the systematic hyperparameter optimization strategy, the metrics for performance and efficiency evaluation, and the computational infrastructure. The central tenet of this methodology is the establishment of a rigorous and equitable comparative framework, achieved by controlling for architectural variations and applying consistent optimization and evaluation protocols across all experiments.

\subsection{Ensuring Fair Comparisons: Methodological Rigor}
\label{sec:methodology_fairness_rigor}

A cornerstone of this paper is the adherence to a strict methodological protocol designed to ensure that all comparisons between the alternative algorithms and backpropagation (BP) are equitable. To address the need for fair benchmarking identified in the literature (see Section~\ref{sec:fair_benchmarking_chap2}), this rigor allows the conclusion that observed differences are primarily due to the training algorithms themselves, rather than to confounding variables. To this end, the author implemented the following principles:

\begin{enumerate}
    \item \textbf{Native Architecture Replication:} Each alternative algorithm was implemented in its specific native architecture, as described or evaluated in its source publication. This principle respects the developmental context of each algorithm. Consequently, FF and MF were evaluated on their designated MLP structures, while CaFo was evaluated on its defined CNN with three blocks for image classification tasks.

    \item \textbf{Identical Architectures for Backpropagation Baselines:} For each experiment involving an alternative algorithm, a corresponding BP baseline was constructed using an \emph{identical} network architecture. This entailed replicating the number and type of layers, neuron and channel counts, activation functions, and any integrated normalization layers. Algorithm-specific components, such as FF's goodness layers or MF's projection matrices, were excluded from the BP baseline, which instead featured a standard final classification layer suitable for end-to-end training. This critical step isolates the impact of the training algorithm from that of the architectural design.

    \item \textbf{Systematic and Universal Hyperparameter Optimization:} To mitigate the influence of default parameters and allow each algorithm to operate near its optimal capacity, systematic hyperparameter optimization was applied using the Optuna framework. This tuning was performed universally for \emph{all} alternative algorithms and their respective BP baselines, for every experimental configuration. This practice ensures that comparisons are not biased by disparities in default tuning.

    \item \textbf{Consistent Early Stopping Based on Validation Performance:} All training procedures incorporated early stopping, governed by performance on a separate validation set. Training was terminated if the validation score failed to improve over a predefined patience period. The model checkpoint that produced the best validation score was subsequently used for final evaluation on the test set and efficiency measurements. This protocol ensures that the results reflect the optimal state achieved by each model and avoids comparisons based on arbitrary training durations, which could unfairly bias algorithms with different convergence dynamics.

    \item \textbf{Consistent Software and Hardware Environment:} All experiments were executed within a unified software environment and on identical hardware (NVIDIA A100 GPUs), thus minimizing the variability due to software versions or hardware idiosyncrasies.

    \item \textbf{Standardized Data Handling and Metrics:} Identical data preprocessing, augmentation, and dataset splits were employed for all compared algorithms within a given task. All performance and efficiency metrics were measured using consistent tools and methodologies in each experiment.
\end{enumerate}
By adhering to these principles, this research provides a robust and transparent evaluation, facilitating reliable conclusions regarding the relative merits and trade-offs of the investigated BP-free training algorithms.

\subsection{Experimental Design Overview}
\label{sec:methodology_design}
The research is based on a comparative experimental design devised by the author to isolate the effect of the training algorithm on both task performance and computational efficiency. The design encompasses the following key comparisons:
\begin{itemize}
    \item \textbf{Forward-Forward (FF) vs. Backpropagation (BP):} A foundational set of experiments was carried out on MLP architectures. This involved implementing FF and a fair BP baseline on a 4$\times$2000 ReLU MLP for the Fashion-MNIST dataset, and on both 3$\times$1000 and 4$\times$2000 ReLU MLPs for the MNIST dataset. These architectures are consistent with those explored in the original FF publication \cite{hinton2022forward} and serve as a reference for BP-free methods.
    \item \textbf{Cascaded-Forward (CaFo) vs. Backpropagation (BP):} A comprehensive evaluation was performed on the MNIST, Fashion-MNIST, CIFAR-10, and CIFAR-100 datasets. CaFo was implemented on its native CNN architecture with three blocks, as specified for image datasets in its source paper \cite{zhao2023cafo}. The evaluation encompassed both the \textbf{CaFo-Rand-CE} and \textbf{CaFo-DFA-CE} variants. Comparisons were made against fair BP baselines utilizing the identical CNN structure with three blocks.
    \item \textbf{Mono-Forward (MF) vs. Backpropagation (BP):} An extensive evaluation of the MF algorithm was performed across the four datasets using its specified native MLP architectures \cite{gong2025mono}. For the MNIST and Fashion-MNIST datasets, a \textbf{2$\times$1000 ReLU MLP} was employed. For the more complex CIFAR-10 and CIFAR-100 datasets, a larger \textbf{3$\times$2000 ReLU MLP} was used. In all instances, MF was benchmarked against fair BP baselines constructed with these identical MLP structures.
\end{itemize}
In every comparison, the principle of using identical architectures for both the alternative algorithm and its corresponding BP baseline was strictly maintained, as detailed in Section \ref{sec:methodology_fairness_rigor}. Furthermore, all training procedures integrated early stopping based on the performance of the validation set. This involved monitoring a specified metric and terminating training if no improvement was observed over a predefined patience period. The final evaluation metrics were reported from the model checkpoint corresponding to the best validation score achieved, a standard practice that ensures that the results reflect the optimal state reached by each model. Although the original publications for FF, CaFo, and MF may have utilized different library versions or fixed training durations, the author conducted all experiments within a consistent, modern software environment (detailed in Section \ref{sec:methodology_infrastructure}) and used early stopping to determine the effective training duration, thus ensuring an equitable comparison of algorithms on contemporary hardware.

\subsection{Dataset Description and Preprocessing}
\label{sec:methodology_datasets}
The following standard benchmark datasets were utilized to evaluate the algorithms across varying levels of classification complexity:
\begin{itemize}
    \item \textbf{MNIST \cite{lecun1998gradient}:} A dataset of 70,000 28x28 pixel grayscale images of handwritten digits (0 to 9), with a standard 60k/10k train/test split.
    \item \textbf{Fashion-MNIST \cite{xiao2017fashion}:} A dataset of 70,000 28x28 pixel grayscale images across 10 categories of apparel, with a standard 60k/10k train/test split.
    \item \textbf{CIFAR-10 \cite{krizhevsky2012learning}:} A dataset of 60,000 32x32 pixel color images spanning 10 object classes, with a standard 50k/10k train/test split.
    \item \textbf{CIFAR-100 \cite{krizhevsky2012learning}:} A dataset similar to CIFAR-10 but with 100 distinct classes, also using a standard 50k/10k train/test split.
\end{itemize}

For each dataset, the standard training and testing splits were preserved. For CIFAR-10, CIFAR-100, and Fashion-MNIST, a validation subset was created by randomly sampling 10\% of the original training data. Following the precedent established in the original FF paper \cite{hinton2022forward}, a fixed split was used for the MNIST dataset, allocating 50,000 images for training and 10,000 for validation. The remainder of the original training set was used for model training in all cases. In adherence to recommended practices for computational workloads on the Athena cluster, datasets were loaded from and intermediate results were stored on the user's high-performance \code{\$SCRATCH} directory on the Lustre filesystem.

Consistent preprocessing steps were applied across all experiments for any given dataset:
\begin{itemize}
    \item \textbf{Normalization:} Pixel values were first scaled to a range via \code{ToTensor}. Subsequently, the images were normalized using the standard means and standard deviations per channel for each dataset:
        \begin{itemize}
            \item \textbf{MNIST:} mean=(0.1307,), std=(0.3081,)
            \item \textbf{Fashion-MNIST:} mean=(0.2860,), std=(0.3530,)
            \item \textbf{CIFAR-10:} mean=(0.4914, 0.4822, 0.4465), std=(0.2023, 0.1994, 0.2010)
            \item \textbf{CIFAR-100:} mean=(0.5071, 0.4867, 0.4408), std=(0.2675, 0.2565, 0.2761)
        \end{itemize}
    \item \textbf{Data Augmentation (CIFAR):} For the CIFAR-10 and CIFAR-100 datasets, standard data augmentation techniques, specifically random horizontal flips (p=0.5) and random crops (32x32 with padding of 4 pixels), were applied consistently during the training of both alternative algorithms and their BP baselines.
    \item \textbf{Data Augmentation (MNIST/Fashion-MNIST):} In accordance with common practice, no data augmentation was applied to the MNIST or Fashion-MNIST datasets.
    \item \textbf{Dimensions:} Input images were processed at their native resolutions. These were 28x28 for MNIST and Fashion-MNIST, and 32x32 for CIFAR-10 and CIFAR-100. The number of input channels was configured accordingly (1 for grayscale, 3 for color).
\end{itemize}

Data loading was managed with PyTorch's \code{DataLoader} class \cite{paszke2019pytorch}. A consistent batch size of 128 was used for training and evaluation across nearly all experiments, and the FF algorithm constituted the sole exception, for which a batch size of 100 was used to align with its reference implementation.

\subsection{Implementation of Alternative Algorithms}
\label{sec:methodology_alternatives}
Each alternative algorithm was implemented to faithfully represent its original description, with a focus on the specific architectural and procedural details required for the experiments in this paper.

\subsubsection{Forward-Forward (FF) Algorithm}
\label{subsec:methodology_ff}
The author's implementation of the FF algorithm, conceptually detailed in Section~\ref{subsec:ff_algorithm_chap2}, was configured for its native MLP architectures.
\begin{itemize}
    \item \textbf{Architectural Details:}
        \begin{itemize}
            \item \textbf{For Fashion-MNIST:} A 4$\times$2000 ReLU MLP was used.
            \item \textbf{For MNIST:} Both a 3$\times$1000 and a 4$\times$2000 ReLU MLP were implemented.
        \end{itemize}
        In all cases, layer normalization was applied as length normalization ($ \mathbf{x}_l / (||\mathbf{x}_l||_2 + \epsilon) $) after each hidden layer's ReLU activation.
    \item \textbf{Mechanism Implementation:}
        \begin{itemize}
            \item Positive and negative data were created by embedding one-hot-encoded labels into the first 10 pixels of the input images.
            \item The goodness function was implemented as the sum of squared activations.
            \item The local logistic loss used a dynamic threshold equal to the number of neurons in the layer.
            \item Both FF layers and a separate downstream classifier were trained using a shared optimizer (AdamW or SGD), with hyperparameters determined by Optuna.
        \end{itemize}
    \item \textbf{Inference Implementation:} Final predictions were determined by adding goodness scores across all hidden layers for each class label and selecting the class with the highest total score.
\end{itemize}

\subsubsection{Cascaded-Forward (CaFo) Algorithm}
\label{subsec:methodology_cafo}
The CaFo algorithm, whose block-wise training paradigm is described in Section~\ref{subsec:cafo_algorithm_chap2}, was implemented for all datasets using its native CNN architecture.
\begin{itemize}
    \item \textbf{Architectural Details:} A CNN with three blocks was implemented, where each block consisted of a 3x3 Convolution, a ReLU activation, a 2x2 Max Pooling layer, and Batch Normalization. The channel depths were set to 32, 128, and 512 for the respective blocks.
    \item \textbf{Predictor Implementation:} A dedicated predictor, consisting of a flattening operation followed by a single Fully Connected (FC) layer, was attached to each of the three blocks.
    \item \textbf{Training Variants Implemented:} The experimental evaluation focused on two primary variants both using use Cross-Entropy (CE) loss for predictor training:
        \begin{itemize}
            \item \textbf{CaFo-Rand-CE:} The CNN blocks were initialized with Kaiming Uniform and remained frozen during training.
            \item \textbf{CaFo-DFA-CE:} The CNN blocks were first pre-trained using DFA, after which their weights were frozen for the subsequent predictor training phase.
        \end{itemize}
    \item \textbf{Predictor Training Protocol:} Each predictor was trained independently using the Adam optimizer. The training process for each predictor was governed by its own early stopping mechanism based on validation performance, with hyperparameters tuned via Optuna.
    \item \textbf{Inference Implementation:} The final class prediction was generated by summing the logits from all three trained predictors.
\end{itemize}

\subsubsection{Mono-Forward (MF) Algorithm}
\label{subsec:methodology_mf}
The implementation of the MF algorithm was based on its native MLP architectures, as detailed conceptually in Section~\ref{subsec:mf_algorithm_chap2}.
\begin{itemize}
    \item \textbf{Architectural Details:}
        \begin{itemize}
            \item \textbf{For MNIST \& Fashion-MNIST:} A \textbf{2$\times$1000 ReLU MLP}.
            \item \textbf{For CIFAR-10 \& CIFAR-100:} A \textbf{3$\times$2000 ReLU MLP}.
        \end{itemize}
        Each hidden layer was augmented with a learnable projection matrix ($\mathbf{M}_i$), and both the layer weights ($\mathbf{W}_i$) and the projection matrices were initialized using the Kaiming uniform method.
    \item \textbf{Training Protocol:} Training was conducted in a layer-wise fashion. For each layer, its weights and projection matrix were trained jointly using the Adam optimizer to minimize a local Cross-Entropy loss. This local loss was derived from the softmax of the goodness scores, which were calculated as described in Equation~\ref{eq:mf_goodness_chap2}. Early stopping based on local validation loss determined the duration of training for each layer.
{\sloppy
\item \textbf{Inference Implementation:} The final prediction was made using the BP-style method, where only the goodness scores from the final hidden layer were used to generate the classification output.
\par}
\end{itemize}

\subsection{Fair Backpropagation Baseline Setup}
\label{sec:methodology_bp_baseline}
For each alternative algorithm experiment, a dedicated BP baseline was meticulously constructed.

\begin{itemize}
    \item \textbf{Architectural Replication:} For every native architecture detailed in Section \ref{sec:methodology_alternatives}, an identical network structure was created for the BP baseline. This replication included the same number and type of layers, neuron/channel counts, ReLU activation functions, and normalization layers, such as BatchNorm retention in the CaFo CNN baseline.
    \item \textbf{Consistent Initialization:} The weights of all layers were initialized using the same method as their counterparts in the implementation of the alternative algorithm, primarily the Kaiming uniform method.
    \item \textbf{Removal of Algorithm-Specific Components:} All components unique to the alternative algorithms were removed. These included the label embedding mechanism of FF, the intermediate predictors of CaFo, the projection matrices $\mathbf{M}_i$ and local losses of MF, and specialized normalizations such as FF's length normalization.
    \item \textbf{Standard BP Training:} The baselines were trained end-to-end using standard BP \cite{rumelhart1986learning}. A final fully connected layer followed by Softmax activation was used for classification, with a global Cross-Entropy loss function being minimized. The AdamW optimizer \cite{loshchilov2017decoupled} was used for training. Crucially, the BP baseline training also employed the same early stopping mechanism as the alternative algorithms, monitoring validation accuracy and saving the checkpoint with the best performance to ensure a fair comparison based on the optimal performance achieved.
\end{itemize}
This configuration ensures that comparisons effectively isolate the influence of the training algorithm by accounting for the optimal training duration determined through early stopping.

\subsection{Hyperparameter Optimization Strategy}
\label{sec:methodology_hyperparameters}
To facilitate a fair comparison and allow each algorithm to perform effectively, a systematic hyperparameter optimization strategy was employed for all evaluated algorithms, including FF, CaFo, MF, and their corresponding BP baselines. This tuning was executed using the Optuna framework \cite{akiba2019optuna}.

The general approach involved defining a search space for key hyperparameters relevant to each algorithm and utilizing Optuna's Tree-structured Parzen Estimator (TPE) sampler \cite{bergstra2011algorithms} to explore this space over a specified number of trials, typically between 30 and 50. The objective of each trial was to maximize the model's accuracy on the separate validation set. The training process within each Optuna trial was subject to the same early stopping protocol used in the final experimental runs. The optimal hyperparameters identified for each configuration were then used for the final training runs reported in Chapter~\ref{chap:experiments}. This consistent tuning methodology ensures a more robust comparison by mitigating the influence of suboptimal default parameters. For efficiency, detailed resource monitoring was generally disabled during the Optuna trials.

\subsubsection{Alternative Algorithm Tuning}
\label{subsec:methodology_hyper_alt}
Dedicated Optuna objective functions were developed to tune each alternative algorithm. Key tuned parameters included:
\begin{itemize}
    \item \textbf{Forward-Forward (FF):} Learning rates and weight decay values for both FF layer updates and separate downstream classifier updates.
    \item \textbf{Cascaded-Forward (CaFo):} Predictor learning rate, weight decay, and the number of training epochs per predictor. For the \textbf{CaFo-DFA} variant, the block learning rate, weight decay, and the number of block training epochs were also tuned.
    \item \textbf{Mono-Forward (MF):} The primary learning rate for updating both layer weights and projection matrices, and possibly the number of training epochs applied per layer.
\end{itemize}
For these algorithms, Optuna's pruner was generally disabled, as their layer-wise or block-wise training dynamics are not consistently amenable to standard pruning mechanisms. Appendix~\ref{app:hyper_tuning} documents the best hyperparameters found for each configuration.

\subsubsection{Backpropagation Baseline Tuning}
\label{subsec:methodology_hyper_bp}
The Optuna framework was also systematically applied to tune the hyperparameters for every BP baseline.
\begin{itemize}
    \item \textbf{Optimizer:} The AdamW optimizer \cite{loshchilov2017decoupled} was used.
    \item \textbf{Search Space:} The learning rate and weight decay were tuned over log-uniform distributions.
    \item \textbf{Objective:} The objective was to maximize validation set accuracy, with the training process within each trial also incorporating early stopping.
    \item \textbf{Procedure:} Optuna was executed for a set number of trials for each BP baseline configuration. A Median pruner was occasionally employed to discard unpromising trials early. The hyperparameters with the best performance were used for the final BP baseline runs, with results documented in Appendix~\ref{app:hyper_tuning}.
\end{itemize}
This process ensures that the BP baselines operate near their optimal performance level on the specific replicated architectures, thereby allowing a fair assessment of the alternative algorithms.

\subsubsection{Rationale for Universal Tuning}
\label{subsec:methodology_hyper_justification}
The decision was made to apply systematic hyperparameter optimization to \textbf{all} training algorithms to ensure the most rigorous and fair comparison possible. Although alternative algorithms are sometimes presented with specific default parameters, these may not be optimal for different datasets, architectures, or the specific training procedures employed in this work. This universal tuning aimed to:
\begin{itemize}
    \item \textbf{Ensure Fairness:} Prevent a situation where an algorithm underperforms merely due to suboptimal hyperparameters.
    \item \textbf{Assess Potential:} Evaluate each algorithm closer to its performance limit as determined by validation accuracy in conjunction with early stopping.
    \item \textbf{Isolate Algorithmic Effects:} More effectively isolate the impact of the core training mechanism from the confounding factor of hyperparameter choice.
\end{itemize}
This universal tuning approach was deemed essential to draw robust conclusions about the relative trade-offs of these novel training paradigms. The resulting hyperparameters are documented in Appendix~\ref{app:hyper_tuning}.

\subsubsection{A Note on "Effective Epochs" and Training Duration}
\label{subsec:methodology_effective_epochs}
A direct consequence of applying early stopping in both end-to-end and layer-wise training paradigms is that the total training effort, measured in epochs, is not directly comparable. For this reason, the term "Effective Epochs" is introduced to quantify this effort, with the following distinct interpretations:
\begin{itemize}
    \item \textbf{For Backpropagation (BP) and Forward-Forward (FF):} Effective Epochs signifieS the total number of \emph{global training epochs} completed for the entire network before the early stopping criterion was met.
    \item \textbf{For Mono-Forward (MF):} Effective Epochs represent the \emph{sum of the actual number of epochs for which each individual layer was trained}, subject to early stopping based on its own local validation loss.
    \item \textbf{For Cascaded-Forward (CaFo):} Effective Epochs are the \emph{sum of the actual epochs for which each predictor was trained} (in the CaFo-Rand-CE variant), or the sum of epochs for block training plus the sum of epochs for predictor training (in the CaFo-DFA-CE variant).
\end{itemize}
This distinction is crucial for an accurate interpretation of the convergence behavior and training times presented in Chapter \ref{chap:experiments}.

\subsection{Metrics for Evaluation}
\label{sec:methodology_metrics}
A comprehensive set of metrics was used to evaluate both performance of the model and computational efficiency:

\begin{itemize}
\item \textbf{Performance Metrics:}
\begin{itemize}
\item \textbf{Classification Accuracy:} The percentage of correct classifications on the test set, calculated using the model checkpoint that achieved the \textbf{best validation accuracy} during training, determined by the early stopping procedure.
\item \textbf{Convergence Rate:} Plotted as accuracy or loss against wall-clock time to visualize the training process until its termination.
\item \textbf{Loss Curves:} Plots of training and validation loss versus epochs.
\end{itemize}
\item \textbf{Efficiency Metrics:}
\begin{itemize}
\item \textbf{Energy Consumption:} The total GPU energy consumed during the main training loop, measured from the start of training \textbf{up to the point of termination}. This was measured in Watt hours (Wh) using the NVML API by integrating the power draw over time.
\item \textbf{Training Time:} The wall-clock time for the main training loop \textbf{until termination}, reported in seconds (s).
\item \textbf{Peak Memory Usage:} The maximum allocated GPU memory during the effective training duration, measured in Mebibytes (MiB) using NVML.
\item \textbf{Floating Point Operations (FLOPs):} Estimated using PyTorch's profiler:
    \begin{itemize}
        \item \textbf{Estimated Forward Pass GFLOPs ($F_{fwd}$):} The GFLOPs required for a single forward pass on one sample, used to quantify inference complexity.
        \item \textbf{Estimated BP Update Cycle GFLOPs ($F_{BP\_update}$):} Calculated \textit{exclusively} for BP baselines to estimate the cost of a full update cycle, a using the heuristic detailed in Section~\ref{sec:energy_efficiency_chap2} (Equation~\ref{eq:bp_update_cost}).
    \end{itemize}
\item \textbf{Estimated Carbon Emissions (gCO2e):} The estimated CO2e generated during the effective training duration. These were calculated using the CodeCarbon library in \textbf{offline mode}, with a specified grid carbon intensity factor corresponding to the hardware's location (Poland: "POL").
\end{itemize}
\end{itemize}
These metrics provide a multifaceted view for analyzing the energy performance trade-offs that result from training procedures optimized via early stopping.

\subsection{Experimental Infrastructure and Tools}
\label{sec:methodology_infrastructure}
All experiments were performed on the \textbf{Athena cluster}, a high-performance computing (HPC) resource provided by ACK Cyfronet AGH.
\begin{itemize}
    \item \textbf{Hardware:} Computations were conducted on the \code{plgrid-gpu-a100} partition. Each experimental run was allocated a \textbf{single NVIDIA A100-SXM4 40GB GPU}. The cluster nodes are equipped with AMD EPYC 7742 CPUs.
    \item \textbf{Job Scheduling:} Jobs were managed via the \textbf{Slurm Workload Manager} and submitted to the \code{plgrid-gpu-a100} partition under the \code{plgoncotherapy-gpu-a100} grant.
    \item \textbf{Configuration Management:} A structured system of \textbf{YAML files} was used, where configurations for each experiment were inherited from a base file defining common defaults.
    \item \textbf{Deep Learning Framework and Environment:} The software environment was built upon loaded modules for \textbf{\code{Python/3.10.4}} and \textbf{\code{CUDA/12.4.0}}. The key libraries installed within a virtual environment included \textbf{PyTorch v2.4.0}, \textbf{Optuna v4.2.1}, \textbf{CodeCarbon v3.0.1}, and \code{pynvml} v12.0.0.
    \item \textbf{Energy/Resource Monitoring:} NVML was utilized to monitor GPU energy consumption, power consumption, and maximum memory usage.
    \item \textbf{FLOPs Profiling:} Forward pass FLOPs were estimated with \textbf{\texttt{torch.profiler}}.
    \item \textbf{Carbon Footprint Estimation:} The CodeCarbon library was used in its offline mode with the ISO code "POL" to estimate CO2e emissions.
    \item \textbf{Experiment Tracking:} Weights \& Biases (\code{wandb}) served as the primary tool to log metrics and track results.
    \item \textbf{Version Control:} Git and GitHub were used for version control. The code repository is detailed in Appendix \ref{app:code_repository}.
\end{itemize}
Detailed environment specifications are provided in Appendix \ref{app:environment}.
\section{Experimental Results}
\label{chap:experiments}
This chapter presents the empirical findings from the comparative experiments conducted according to the fair benchmarking framework and methodology detailed in Chapter~\ref{chap:methodology}. The results focus on evaluating the performance and energy efficiency of the alternative algorithms against their respective fair backpropagation (BP) baselines across the MNIST, Fashion-MNIST, CIFAR-10, and CIFAR-100 datasets.

Although performance is primarily assessed through the accuracy of classification and convergence behavior, efficiency evaluation utilizes a multifaceted approach. To provide a robust assessment of practical computational and energy efficiency, the analysis prioritizes directly measured hardware metrics. These primary indicators include total energy consumption, measured in Watt-hours (Wh) using the NVML API, the wall-clock training time until termination by early stopping, and the peak GPU memory usage, also monitored via NVML. These measurements reflect the utilization of real-world resources on the specified NVIDIA A100 hardware platform. To supplement these hardware-specific metrics, estimated GFLOPs for a single forward pass ($F_{fwd}$), obtained using the PyTorch Profiler, are reported to provide a hardware-agnostic measure of arithmetic complexity. For BP baselines, the estimated GFLOPs per update cycle ($F_{BP\_update} \approx 3 \times F_{fwd}$) are included to illustrate the theoretical computational difference that arises from the backward pass. Furthermore, an estimated CO2e is reported, calculated to provide an environmental impact perspective based on the measured energy consumption within the specified compute region (Poland, ISO code: POL). The subsequent sections detail the results for each algorithm and its corresponding baseline, followed by comparative analyzes focusing on these key performance and efficiency indicators.

\subsection{Forward-Forward (FF) vs. Backpropagation (BP) Comparison}
\label{sec:exp_ff_vs_bp}
This section presents a foundational analysis comparing Geoffrey Hinton's FF algorithm against a fair BP baseline, focusing on the MLP architectures for which FF was primarily designed. The initial performance results reveal a significant paradox: while FF can achieve impressive accuracy, thus validating its core concept, it does so at a drastically higher convergence cost.

\begin{table}[htbp]
    \centering
    \caption{Performance Comparison: FF vs. BP on MLP Architectures (Mean ± Std Dev over 3 runs).}
    \label{tab:ff_bp_mlp_perf_results_ch4}
    \begin{tabular}{lll c c}
        \toprule
        Dataset & Architecture & Algorithm & Test Accuracy (\%) & Effective Epochs \\
        \midrule
        \multirow{2}{*}{F-MNIST} & \multirow{2}{*}{4 $\times$ 2000 MLP} & FF-AdamW & \textbf{89.63 ± 0.12} & 90.33 ± 16.86 \\
         & & BP Baseline& 88.88 ± 0.61 & \textbf{15.67 ± 0.58} \\
        \midrule
        \multirow{3}{*}{MNIST} & \multirow{3}{*}{3 $\times$ 1000 MLP} & FF-AdamW & \textbf{98.33 ± 0.03} & 77.00 ± 0.00 \\
         & & FF-SGD & 98.23 ± 0.05 & 99.33 ± 1.15 \\
         & & BP Baseline & 98.09 ± 0.06 & \textbf{35.00 ± 0.00} \\
        \midrule
        \multirow{2}{*}{MNIST} & \multirow{2}{*}{4 $\times$ 2000 MLP} & FF-AdamW & 98.42 ± 0.14 & 87.00 ± 11.53 \\
         & & BP Baseline & \textbf{98.43 ± 0.03} & \textbf{49.00 ± 0.00} \\
        \bottomrule
    \end{tabular}
    \caption*{\footnotesize Effective Epochs: Average number of global training epochs completed before early stopping. Best results in each comparison group are highlighted in \textbf{bold}. For accuracy, higher is better; for epochs, lower is better.}
\end{table}

As shown in Table~\ref{tab:ff_bp_mlp_perf_results_ch4}, the FF algorithm successfully trains MLP architectures to a final test accuracy that is competitive with, and in some cases slightly superior to, a fairly tuned BP baseline on both the MNIST and Fashion-MNIST datasets.  This result is significant, as it validates the core premise of Hinton's work: effective learning is indeed possible without propagating error gradients backwards.

However, the Effective Epochs metric in Table~\ref{tab:ff_bp_mlp_perf_results_ch4} reveals the steep cost of this accuracy. FF consistently required a dramatically higher number of training epochs to reach its optimal validation checkpoint, often more than double that of BP, indicating a far less efficient learning process. The practical implications of this inefficiency are best understood by analyzing the convergence behavior with respect to the actual wall-clock time.

\begin{figure}[htbp] 
    \centering
    \begin{subfigure}{0.85\textwidth}
        \centering
        \includegraphics[width=\linewidth]{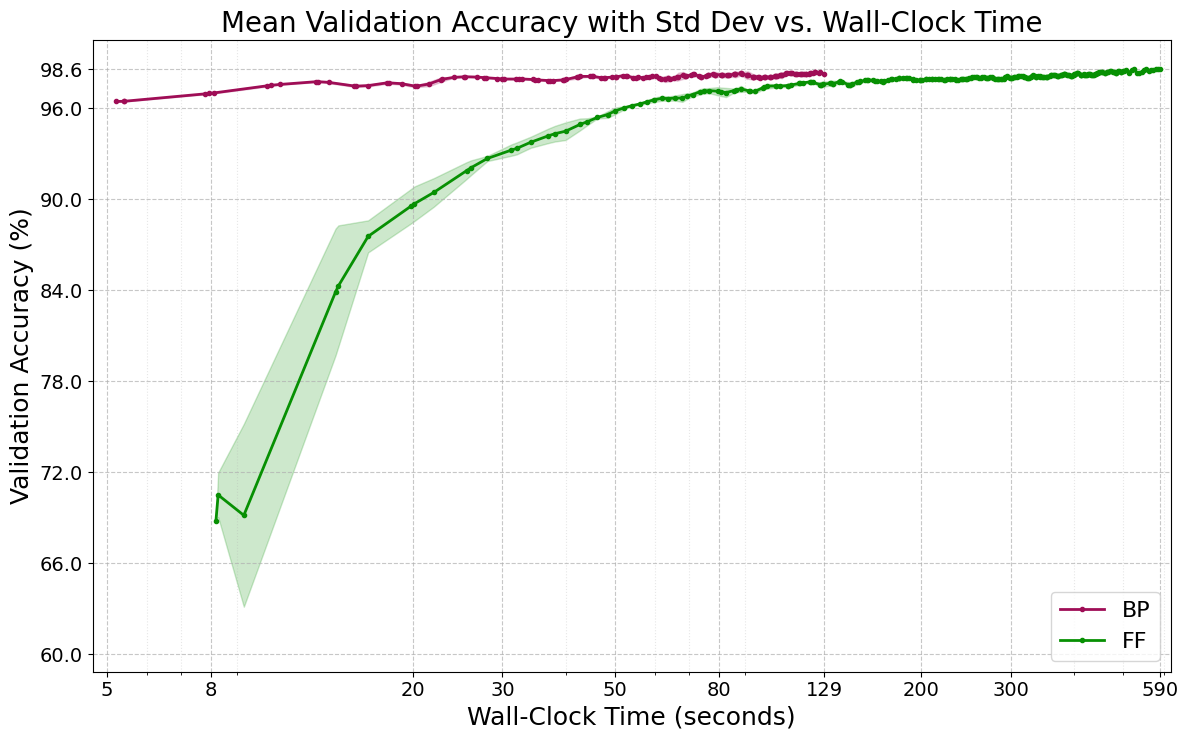}
        \caption{}
        \label{fig:ch4_ff_bp_mnist_mlp_4x2000_val_acc_time}
    \end{subfigure}
    \hfill
    \begin{subfigure}{0.85\textwidth}
        \centering
        \includegraphics[width=\linewidth]{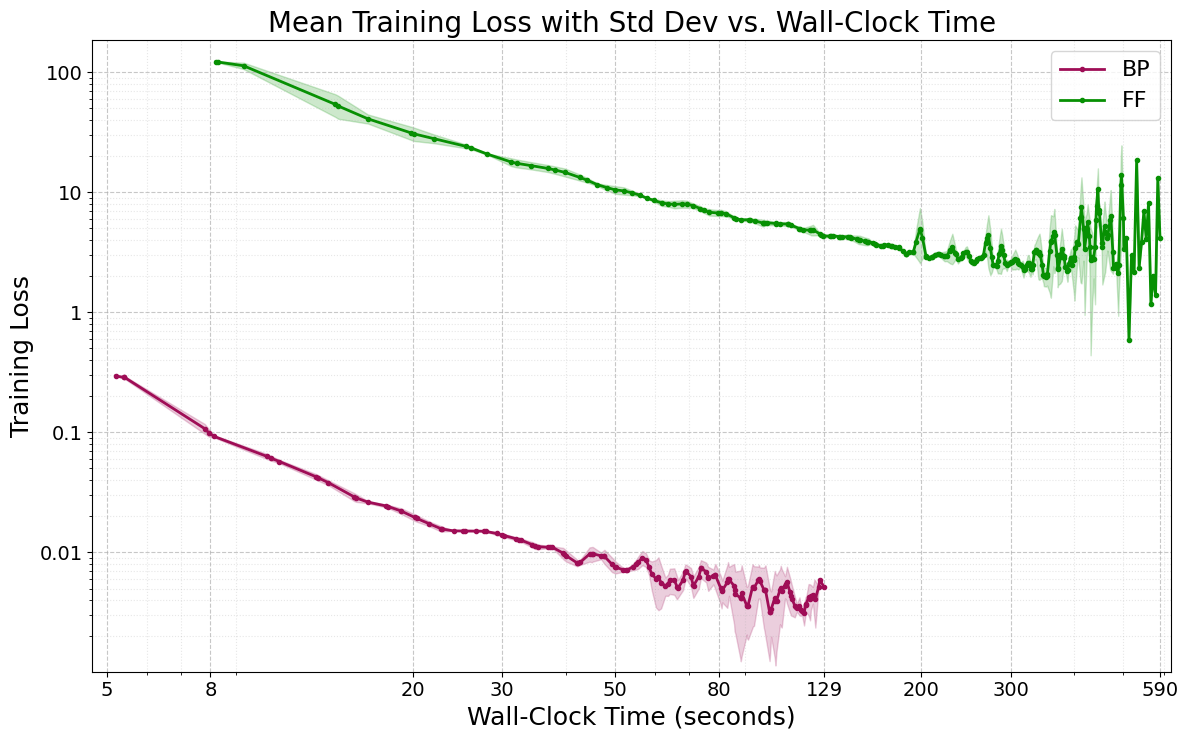}
        \caption{}
        \label{fig:ch4_ff_bp_mnist_mlp_4x2000_train_loss_time}
    \end{subfigure}
    \caption{Convergence Dynamics for FF vs. BP on MNIST 4 $\times$ 2000 MLP.}
    \label{fig:ch4_ff_bp_mnist_mlp_4x2000_conv_time}
\end{figure}

Figure~\ref{fig:ch4_ff_bp_mnist_mlp_4x2000_conv_time} provides a clear representation of this inefficiency. The validation accuracy plot (Figure~\ref{fig:ch4_ff_bp_mnist_mlp_4x2000_val_acc_time}) shows that the BP baseline achieves near-optimal performance rapidly, in well under 100 seconds, with a stable and consistent learning trajectory. In stark contrast, FF embarks on a slow, gradual, and more volatile learning path, which requires more than 400 seconds to reach a similar accuracy level. The training loss plot (Figure~\ref{fig:ch4_ff_bp_mnist_mlp_4x2000_train_loss_time}) reveals the underlying algorithmic cause for this disparity: BP's loss decreases smoothly and efficiently, whereas FF's loss decreases slowly and erratically from a much higher starting point, ultimately plateauing at a level orders of magnitude above that of BP. This indicates that FF's collection of local "goodness" objectives serves as a far weaker and less direct proxy for the global classification task, resulting in a fundamentally less efficient optimization landscape that necessitates a protracted training process with severe consequences for overall resource consumption.

The slow convergence of the FF algorithm directly translates into a significant increase in computational cost across all measured efficiency metrics. As detailed in Table~\ref{tab:ff_bp_mlp_eff_results_ch4}, the training time required to reach the best model was substantially longer (a 4x to 13x increase) and the total energy consumption was significantly higher (a 3x to 10x increase) for FF compared to BP across all configurations. This led to a correspondingly larger estimated CO2e footprint. 

\begin{table}[htbp]
    \centering
    \caption{Efficiency Comparison: FF vs. BP on MLP Architectures (Mean ± Std Dev over 3 runs).}
    \label{tab:ff_bp_mlp_eff_results_ch4}
    \resizebox{\textwidth}{!}{%
    \begin{tabular}{ccc c@{ ± }c c@{ ± }c c@{ ± }c c c}
        \toprule
        Dataset & Architecture & Algorithm & \multicolumn{2}{c}{Train Time (s)} & \multicolumn{2}{c}{Energy (Wh)} & \multicolumn{2}{c}{Est. CO2e (g)} & Peak Mem (MiB) & Total GFLOPs \\
        \midrule
        \multirow{2}{*}{F-MNIST} & \multirow{2}{*}{4 $\times$ 2000 MLP} & FF-AdamW & 574.60 & 134.49 & 14.28 & 2.66 & 23.10 & 8.52 & 1190 & \textbf{0.027} \\
        & & BP Baseline & \textbf{43.09} & \textbf{0.80} & \textbf{1.48} & \textbf{0.01} & \textbf{1.67} & \textbf{0.01} & \textbf{1168} & 0.082 \\
        \midrule
        \multirow{3}{*}{MNIST} & \multirow{3}{*}{3 $\times$ 1000 MLP} & FF-AdamW & 370.72 & 8.63 & 6.68 & 0.30 & 10.44 & 0.82 & 954 & \textbf{0.006} \\
        & & FF-SGD & 444.15 & 5.07 & 7.89 & 0.16 & 12.26 & 0.73 & 956 & \textbf{0.006} \\
        & & BP Baseline& \textbf{83.85} & \textbf{1.12} & \textbf{1.52} & \textbf{0.01} & \textbf{2.55} & \textbf{0.29} & \textbf{946} & 0.017 \\
        \midrule
        \multirow{2}{*}{MNIST} & \multirow{2}{*}{4 $\times$ 2000 MLP} & FF-AdamW & 509.59 & 73.20 & 14.26 & 1.74 & 17.50 & 2.74 & 1190 & \textbf{0.027} \\
        & & BP Baseline & \textbf{125.80} & \textbf{3.13} & \textbf{4.23} & \textbf{0.11} & \textbf{5.25} & \textbf{0.07} & \textbf{1168} & 0.082 \\
        \bottomrule
    \end{tabular}%
    }
    \caption*{\footnotesize Total GFLOPs: Estimated GigaFLOPs per sample. For FF, this is the cost of a single forward pass ($F_{fwd}$). For BP, it is the estimated cost of a full update cycle ($F_{BP\_update}$). Best efficiency results in each comparison group are highlighted in \textbf{bold} (lower is better).}
\end{table}

\begin{figure}[htbp]
    \centering
    \begin{subfigure}{1.0\textwidth}
        \centering
        \includegraphics[width=\linewidth]{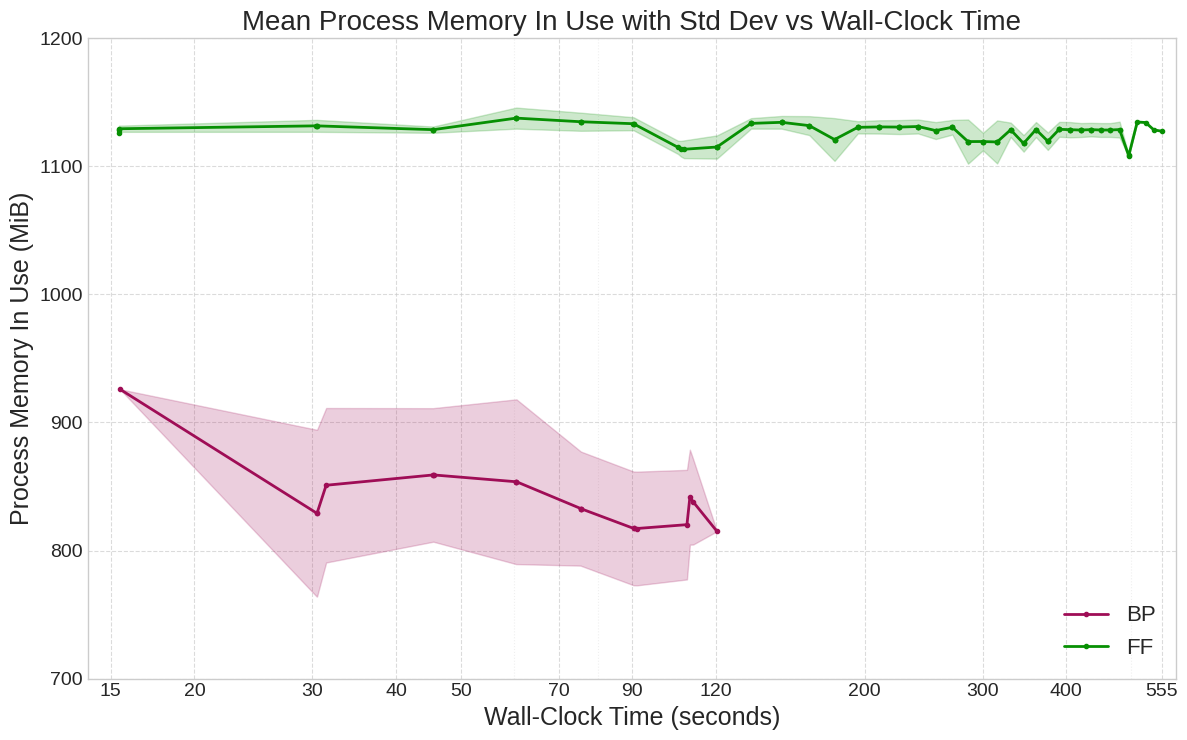}
        \caption{}
        \label{fig:ch4_ff_bp_mnist_mlp_4x2000_proc_mem_time}
    \end{subfigure}
    \hfill
    \begin{subfigure}{1.0\textwidth}
        \centering
        \includegraphics[width=\linewidth]{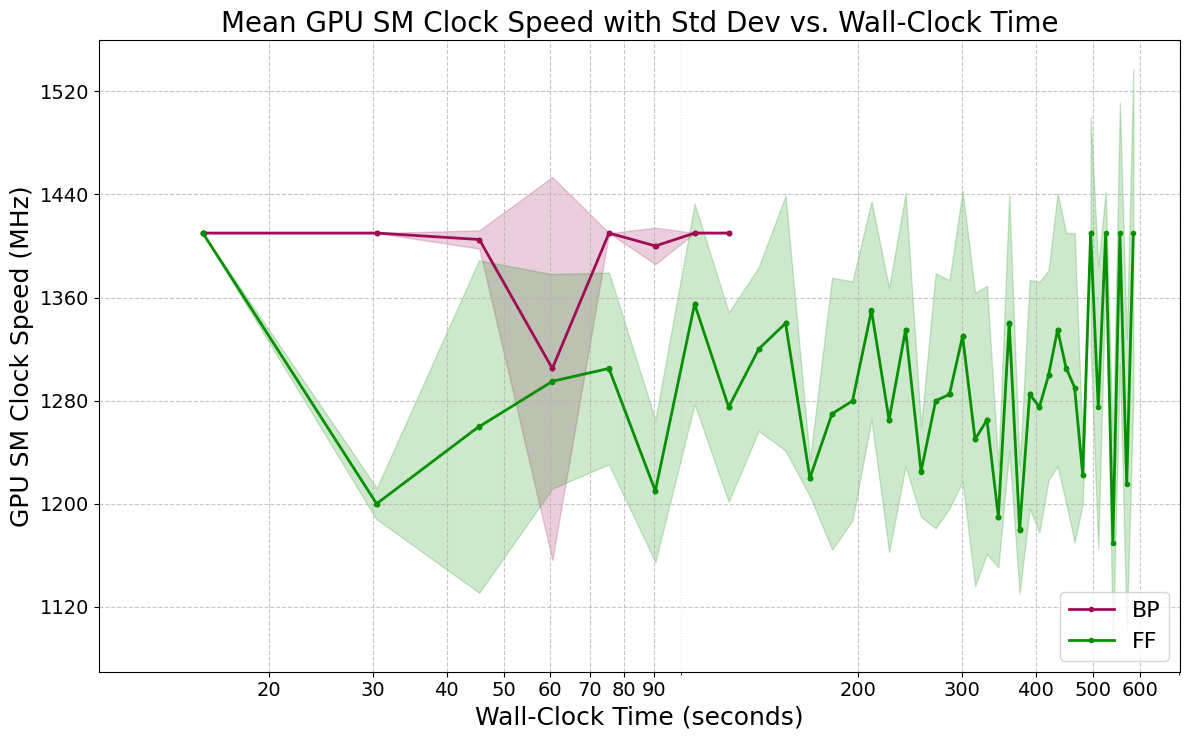}
        \caption{}
        \label{fig:ch4_ff_bp_mnist_mlp_4x2000_sm_clock_time}
    \end{subfigure}
    \caption{GPU Resource Utilization Profiles for FF vs. BP on MNIST 4 $\times$ 2000 MLP.}
    \label{fig:ch4_ff_bp_mnist_mlp_4x2000_resource_util}
\end{figure}

A definitive explanation for this inefficiency emerges from an analysis of the hardware-level monitoring plots in Figure~\ref{fig:ch4_ff_bp_mnist_mlp_4x2000_resource_util}. The memory plot (Figure~\ref{fig:ch4_ff_bp_mnist_mlp_4x2000_proc_mem_time}) reveals a critical and counterintuitive finding: contrary to the theoretical benefit of avoiding activation storage, \textbf{the FF algorithm consistently used slightly more GPU peak memory than BP in these experiments}. This observation empirically refutes a key presumed advantage, suggesting that practical implementation overheads, such as managing separate local optimizers, handling positive and negative data streams, or other specific PyTorch implementation details, impose a memory cost that outweighs the theoretical savings for these MLPs. More revealing is the GPU SM clock speed plot (Figure~\ref{fig:ch4_ff_bp_mnist_mlp_4x2000_sm_clock_time}), which provides a clear indicator of FF's inefficiency. While BP maintains a high, stable GPU clock speed, indicating that it effectively saturates the hardware with large, parallelizable matrix multiplications, FF's clock speed is erratic, "spiky," and operates at a lower average. This pattern signifies a computational bottleneck in which the GPU is frequently stalled, likely due to a high volume of small, sequential operations that lead to poor hardware utilization. Consequently, \textbf{FF's slowness is not due to its computations being inherently more complex, but to its failure to efficiently utilize modern, highly parallel hardware}.

\textbf{In summary}, the FF algorithm serves as a vital foundational exploration, confirming its conceptual validity by training MLP architectures to an accuracy competitive with backpropagation. However, this success comes at a severe and disproportionate efficiency cost. The combination of a less effective optimization process and inefficient hardware utilization results in drastically longer training times, higher energy consumption, and a larger estimated CO2e footprint, while theoretical memory savings were not realized in this practical context. These findings quantify FF's shortcomings and motivate the investigation of more structured approaches like CaFo, which aim to resolve these inefficiencies by integrating supervisory signals more directly.

\subsection{Cascaded-Forward (CaFo) vs. Backpropagation (BP) Comparison}
\label{sec:exp_cafo_vs_bp_ch4}

This section presents a holistic analysis of the CaFo algorithm on its native 3-block CNN architecture, moving beyond foundational concepts to evaluate its practical viability as a structured BP-free alternative. The investigation covers two primary variants, CaFo-Rand-CE (using randomly initialized, fixed blocks) and CaFo-DFA-CE (using blocks pre-trained with DFA), against a fair BP baseline. By integrating dynamic convergence behavior with final, static performance, and efficiency metrics, a complete picture of the algorithm's inherent trade-offs is constructed. To understand CaFo's practical efficiency, it is crucial to first analyze the distinct convergence behaviors of its different variants.

\begin{figure}[htbp]
    \centering
    \begin{subfigure}{0.80\textwidth}
        \centering
        \includegraphics[width=\linewidth]{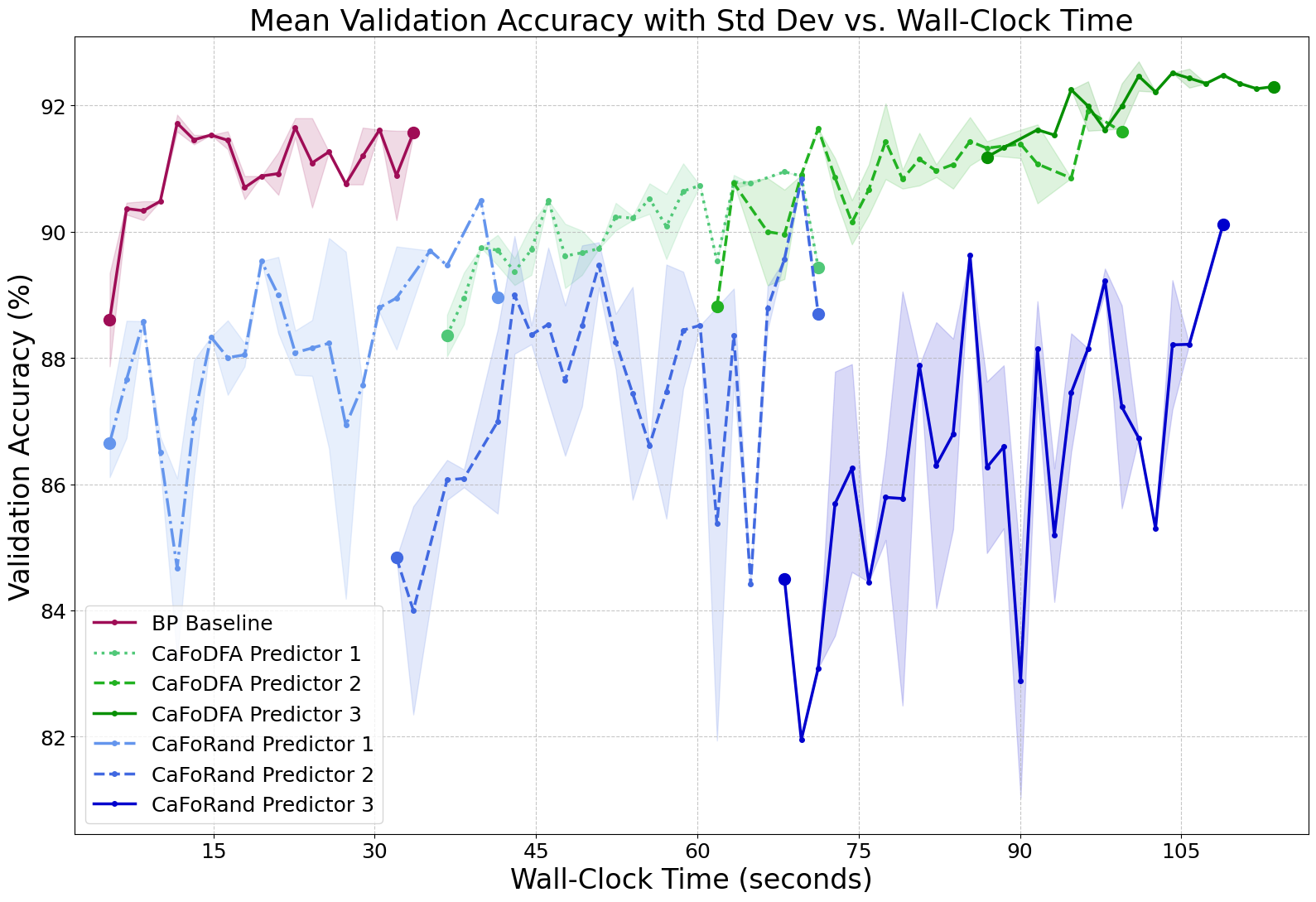}
        \label{fig:ch4_cafo_bp_fmnist_cnn_val_acc_time_reordered}
    \end{subfigure}
    \hfill
    \begin{subfigure}{0.80\textwidth}
        \centering
        \includegraphics[width=\linewidth]{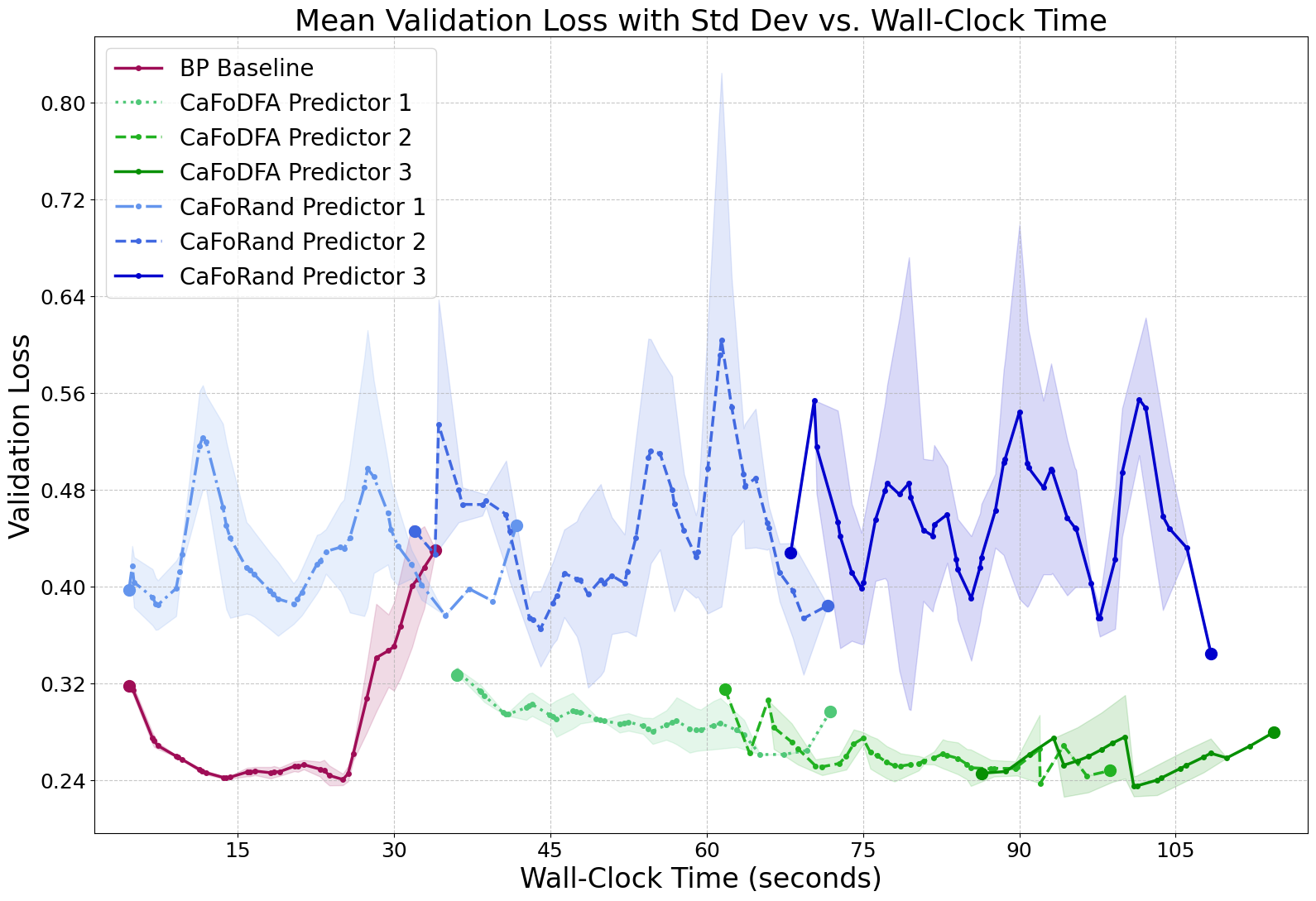}
        \label{fig:ch4_cafo_bp_fmnist_cnn_val_loss_time_reordered}
    \end{subfigure}
    \caption{Convergence Dynamics for CaFo Variants and BP on Fashion-MNIST 3-block CNN.}
    \label{fig:ch4_cafo_bp_fmnist_cnn_conv_curves_reordered}
\end{figure}

The convergence plots in Figure~\ref{fig:ch4_cafo_bp_fmnist_cnn_conv_curves_reordered} offer a granular view of the learning process. The BP baseline exemplifies an ideal learning trajectory, exhibiting rapid, smooth, and stable convergence to a high-accuracy plateau, showcasing the efficacy of end-to-end optimization guided by a global error signal. In contrast, the CaFo-DFA-CE variant demonstrates a clear two-phase process: an initial period of no predictor activity represents the upfront computational cost of pre-training the feature extraction blocks via DFA, after which the individual predictors begin training sequentially and learn quickly, with the predictor leveraging the deepest features ultimately achieving the highest accuracy. However, the CaFo-Rand-CE variant exhibits a highly volatile and inefficient learning process, with accuracy fluctuating wildly, indicating that the untrained random features are insufficient for stable credit assignment. The loss curves corroborate these findings, with BP’s loss decreasing rapidly to a low, stable minimum, while the loss for CaFo-Rand-CE remains high and erratic. The crucial insight from this dynamic analysis is the trade-off between feature quality and computational cost: the superior performance of CaFo-DFA comes at the price of a significant upfront time and energy investment in block training. The failure of the CaFo-Rand variant underscores that effective BP-free learning cannot be achieved by simply attaching local classifiers to random projections; the quality of intermediate representations is paramount.

The dynamic behaviors are quantified in the final performance and efficiency metrics, which provide a static comparison of the final outcomes and allow a holistic assessment.
\begin{table}[t!] 
    \centering
    \caption{Performance Comparison: CaFo Variants vs. BP on Native 3-block CNN (Mean ± Std Dev over 3 runs).}
    \label{tab:cafo_bp_cnn_perf_results_ch4_reordered}
    \resizebox{\textwidth}{!}{%
    \begin{tabular}{ccccc}
        \toprule
        Dataset & Architecture & Algorithm & Test Accuracy (\%) & Total Effective Epochs \\
        \midrule
        \multirow{3}{*}{MNIST} & \multirow{3}{*}{3-block CNN} & CaFo-Rand-CE   & 98.62 ± 0.02             & 30.00 ± 0.00                     \\
                               & & CaFo-DFA-CE    & \textbf{99.02 ± 0.05}             & 37.00 ± 1.00   \\ 
                               & & BP Baseline    & 98.94 ± 0.15             & \textbf{15.33 ± 0.58}                  \\
        \midrule
        \multirow{3}{*}{F-MNIST} & \multirow{3}{*}{3-block CNN} & CaFo-Rand-CE   & 90.36 ± 0.45             & 44.33 ± 2.08                     \\
                                       & & CaFo-DFA-CE    & \textbf{91.72 ± 0.12}             & 47.33 ± 3.51  \\ 
                                       & & BP Baseline    & 89.25 ± 0.34             & \textbf{13.33 ± 0.58}                  \\
        \midrule
        \multirow{3}{*}{CIFAR-10} & \multirow{3}{*}{3-block CNN} & CaFo-Rand-CE  & 66.09 ± 0.32             & 72.67 ± 15.01                  \\
                                  & & CaFo-DFA-CE    & 77.60 ± 0.25             & 295.00 ± 6.56 \\ 
                                  & & BP Baseline    & \textbf{79.32 ± 3.99}             & \textbf{71.67 ± 3.79}                  \\
        \midrule
        \multirow{3}{*}{CIFAR-100} & \multirow{3}{*}{3-block CNN} & CaFo-Rand-CE  & 41.19 ± 0.29             & 136.33 ± 9.81                 \\
                                   & & CaFo-DFA-CE    & 48.20 ± 0.22             & 261.33 ± 11.37 \\ 
                                   & & BP Baseline    & \textbf{52.63 ± 5.20}             & \textbf{37.33 ± 2.52}                  \\
        \bottomrule
    \end{tabular}%
    }
    \caption*{\footnotesize Total Effective Epochs: Sum of actual epochs each component was trained for. Best results in each comparison group are in \textbf{bold}. For accuracy, higher is better; for epochs, lower is better.}
\end{table}

The test accuracy results in Table~\ref{tab:cafo_bp_cnn_perf_results_ch4_reordered} confirm this performance hierarchy, with backpropagation setting a strong benchmark. CaFo-DFA-CE emerges as the most competitive alternative, significantly closing the accuracy gap on complex datasets like CIFAR-10 and even surpassing BP on less complex datasets like Fashion-MNIST. In contrast, CaFo-Rand-CE consistently lags in performance, especially on CIFAR-10 and CIFAR-100, where it suffers a substantial accuracy drop of over 13 and 11 percentage points, respectively. This finding is directly in line with the work of \cite{zhao2023cafo}, who also demonstrated the superiority of using trained blocks.

\begin{table}[htbp]
    \centering
    \caption{Efficiency Comparison: CaFo Variants vs. BP on Native 3-block CNN (Mean ± Std Dev over 3 runs).}
    \label{tab:cafo_bp_cnn_eff_results_ch4_reordered_eff}
    \resizebox{\textwidth}{!}{%
    \begin{tabular}{cc c@{ ± }c c@{ ± }c c@{ ± }c c c}
        \toprule
        Dataset & Algorithm & \multicolumn{2}{c}{Train Time (s)} & \multicolumn{2}{c}{Energy (Wh)} & \multicolumn{2}{c}{Est. CO2e (g)} & Peak Mem (MiB) & Total GFLOPs \\
        \midrule
        \multirow{3}{*}{MNIST} & CaFo-Rand-CE    & 69.20 & 0.69 & 1.20 & 0.02 & 2.06 & 0.25 & \textbf{1008} & \textbf{0.07271} \\
                               & CaFo-DFA-CE     & 84.52 & 2.24 & 1.64 & 0.05 & 2.74 & 0.49 & 1092 & \textbf{0.07271} \\
                               & BP Baseline     & \textbf{37.61} & \textbf{0.18} & \textbf{0.90} & \textbf{0.02} & \textbf{1.33} & \textbf{0.04} & 1080 & 0.21840 \\
        \midrule
        \multirow{3}{*}{F-MNIST} & CaFo-Rand-CE    & 101.09 & 3.55 & 1.76 & 0.07 & 2.80 & 0.05 & \textbf{1008} & \textbf{0.07271} \\
                                       & CaFo-DFA-CE     & 105.37 & 7.40 & 2.48 & 0.11 & 3.51 & 0.27 & 1092 & \textbf{0.07271} \\
                                       & BP Baseline     & \textbf{33.18} & \textbf{0.94} & \textbf{0.81} & \textbf{0.04} & \textbf{1.13} & \textbf{0.06} & 1080 & 0.21840 \\
        \midrule
        \multirow{3}{*}{CIFAR-10} & CaFo-Rand-CE    & \textbf{332.79} & \textbf{61.94} & \textbf{5.50} & \textbf{1.00} & \textbf{10.22} & \textbf{1.10} & \textbf{1014} & \textbf{0.09614} \\
                                  & CaFo-DFA-CE     & 1314.38 & 37.66 & 27.37 & 0.44 & 45.71 & 2.14 & 1128 & \textbf{0.09614} \\
                                  & BP Baseline     & 339.51 & 18.66 & 6.81 & 0.46 & 10.93 & 1.28 & 1114 & 0.28893 \\
        \midrule
        \multirow{3}{*}{CIFAR-100} & CaFo-Rand-CE    & 631.60 & 38.83 & 10.56 & 0.77 & 18.62 & 0.33 & \textbf{1060} & \textbf{0.09614} \\
                                   & CaFo-DFA-CE     & 1161.79 & 60.05 & 24.09 & 1.32 & 39.23 & 4.88 & 1144 & \textbf{0.09614} \\
                                   & BP Baseline     & \textbf{180.57} & \textbf{11.23} & \textbf{3.56} & \textbf{0.17} & \textbf{5.63} & \textbf{0.36} & 1116 & 0.29334 \\
        \bottomrule
    \end{tabular}%
    }
    \caption*{\footnotesize Total GFLOPs: Estimated GigaFLOPs per sample ($F_{fwd}$ for CaFo, $F_{BP\_update}$ for BP). Best efficiency results in each group are in \textbf{bold} (lower is better).}
\end{table}

However, the central trade-off of the CaFo algorithm becomes evident in the efficiency metrics detailed in Table~\ref{tab:cafo_bp_cnn_eff_results_ch4_reordered_eff}. \pagebreak

The results reveal the substantial cost of CaFo-DFA-CE's accuracy; on CIFAR-10, for instance, it required approximately four times more training time and energy and generated four times more estimated CO2e than the BP baseline, all while achieving slightly lower accuracy. This highlights the critical weakness of the CaFo-DFA algorithm: its method for BP-free feature learning, while effective, is computationally expensive. The CaFo-Rand-CE variant, on the other hand, illuminates a potential efficiency niche, proving approximately 19\% more energy efficient than BP on CIFAR-10, but this came at the cost of a 13 percentage point drop in accuracy. 

A particularly nuanced finding emerges from the analysis of peak memory consumption. The CaFo-Rand-CE variant consistently validates the theoretical benefit of BP-free methods, using 5 to 9\% less peak memory than the BP baseline by avoiding the storage of activations for a deep backward pass. However, in a counterintuitive result, the CaFo-DFA-CE variant consistently consumed more peak memory than BP. This empirically demonstrates that the memory overhead of the DFA training mechanism itself, such as storing fixed random feedback matrices, can exceed the savings gained from eliminating the global backward pass, thus refuting any generalized assumption that all BP-free methods are inherently more memory-efficient.

A comparison with the original CaFo paper reveals both consistencies and notable differences attributable to the rigorous methodology employed in this study. The author's implementation of CaFo-DFA-CE achieved notably better accuracy on CIFAR-10 (22.40\% error vs. 30.52\%) and CIFAR-100 (51.80\% error vs. 57.87\%) while also training significantly faster. These improvements are likely attributable to the systematic use of Optuna for hyperparameter tuning and the consistent application of early stopping for each predictor based on its validation loss, which allows each component to train to its optimal point and avoids the fixed, and potentially excessive, epoch counts used in the original work.

In the broader landscape of BP-free learning, \textbf{the CaFo algorithm represents a significant conceptual advance over simpler ideas such as FF}. Its structured block-wise approach allows it to achieve respectable accuracy on complex CNN architectures. Despite these architectural innovations, the empirical results show that CaFo does not present a universally superior alternative to backpropagation. Instead, it presents the practitioner with a stark choice: backpropagation offers the best overall balance of high accuracy and outstanding time and energy efficiency; the CaFo-DFA-CE variant can achieve near-BP accuracy but at a prohibitive computational cost, making it impractical for most energy-constrained applications; and finally, the CaFo-Rand-CE variant offers modest memory and potential energy benefits but with a significant performance penalty on complex tasks. The challenges highlighted by CaFo's trade-off profile provide a strong motivation to investigate other advanced algorithms like MF, which aims to achieve BP-competitive accuracy with a more efficient local learning mechanism, thus setting the stage for the next phase of this investigation.

\subsection{Mono-Forward (MF) vs. Backpropagation (BP) Comparison}
\label{sec:exp_mf_vs_bp_ch4_final}

This section presents a comprehensive empirical evaluation of the MF algorithm, the final and most advanced BP-free method investigated in this paper, focusing on its performance and efficiency in its native MLP architectures. The results robustly demonstrate that for the MLP architectures under investigation, \textbf{MF is not merely a theoretical alternative but a practical, high performance and computationally efficient training algorithm}. It consistently achieves superior accuracy while demanding significantly fewer computational resources, establishing it as a compelling alternative to conventional backpropagation for this class of networks.

The empirical evidence indicates that MF's local, greedy learning approach consistently matches or slightly surpasses the classification accuracy of a fairly tuned BP baseline on identical MLP structures. This performance advantage is most pronounced on the more complex CIFAR-10 task, where MF achieves a notable 1.21 percentage point lead, as detailed in Table~\ref{tab:mf_bp_mlp_perf_ch4}.

\begin{table}[htbp]
    \centering
    \caption{Performance Comparison: MF vs. BP on Native MLP Architectures (Mean ± Std Dev over 3 runs).}
    \label{tab:mf_bp_mlp_perf_ch4}
    \resizebox{\textwidth}{!}{%
    \begin{tabular}{ccccc}
        \toprule
        Dataset        & Architecture     & Algorithm                  & Test Accuracy (\%) & Effective Epochs \\
        \midrule
        \multirow{2}{*}{MNIST} & \multirow{2}{*}{2 $\times$ 1000 MLP} & MF        & \textbf{98.14 ± 0.02}             & \textbf{15.33 ± 0.58} \\
        & & BP & 98.05 ± 0.10             & 16.00 ± 0.00 \\
        \midrule
        \multirow{2}{*}{F-MNIST}  & \multirow{2}{*}{2 $\times$ 1000 MLP}     & MF        & \textbf{89.72 ± 0.11}             & 23.00 ± 0.00 \\
        & & BP & 89.21 ± 0.50             & \textbf{17.67 ± 1.15} \\
        \midrule
        \multirow{2}{*}{CIFAR-10}       & \multirow{2}{*}{3 $\times$ 2000 MLP}     & MF         & \textbf{62.34 ± 0.05}             & \textbf{38.00 ± 0.00} \\
        & & BP & 61.13 ± 0.34             & 50.67 ± 2.08 \\
        \midrule
        \multirow{2}{*}{CIFAR-100}      & \multirow{2}{*}{3 $\times$ 2000 MLP}     & MF         & \textbf{30.31 ± 0.15}             & 24.00 ± 0.00 \\
        & & BP & 29.94 ± 0.37             & \textbf{21.33 ± 2.31} \\
        \bottomrule
    \end{tabular}%
    }
    \caption*{\footnotesize Effective Epochs: Sum of layer-wise epochs for MF, global epochs for BP. Best results in each comparison group are in \textbf{bold}. For accuracy, higher is better; for epochs, lower is better.}
\end{table}

\begin{figure}[t!] 
    \centering
    \includegraphics[width=0.9\textwidth]{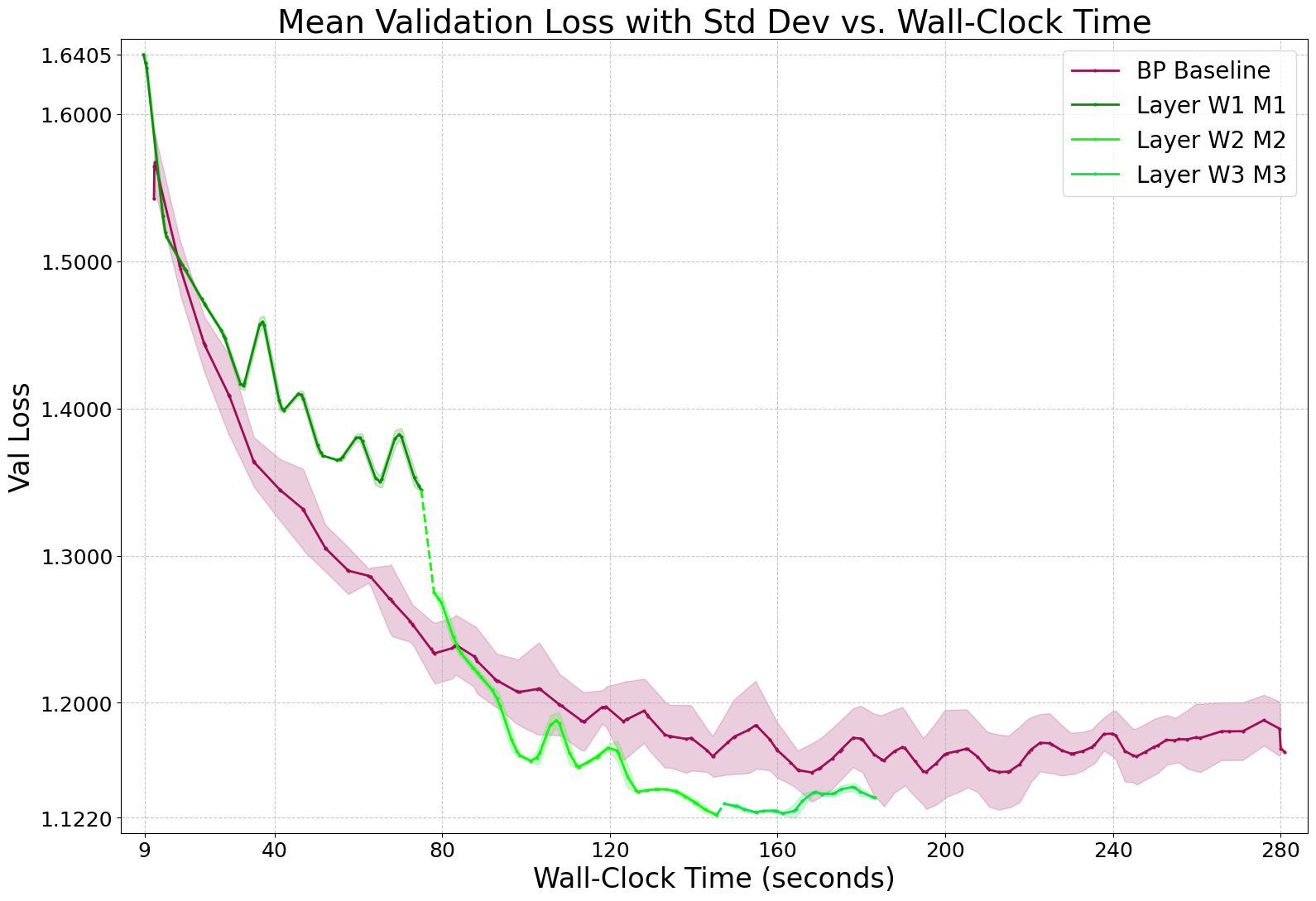} 
    \caption{Mean Validation Loss Dynamics for MF vs. BP on CIFAR-10 3 $\times$ 2000 MLP.}
    \label{fig:mf_bp_cifar10_mlp_conv_curves}
\end{figure}

The key explanation for this superior generalization lies in the dynamics of the validation loss, illustrated in Figure~\ref{fig:mf_bp_cifar10_mlp_conv_curves}, which shows that the sequence of local optimizations performed by MF converges to a more favorable and lower final validation loss compared to the best loss achieved by the end-to-end global optimization of backpropagation. \textbf{This suggests that MF's layer-wise cascade is a more effective optimization strategy for these MLP architectures, leading to better generalization on the test set}.

However, the most significant findings relate to MF's profound computational and energy efficiency gains. The algorithm demonstrates a marked advantage in the time required to reach its optimal validation checkpoint, proving approximately 33.8\% faster than its BP counterpart on the CIFAR-10 MLP. The combination of lower power consumption and shorter training duration results in substantial energy savings. On the CIFAR-10 task, MF consumed approximately 40.8\% less energy and produced 39.2\% less estimated CO2e than BP did, highlighting its potential, as quantified in Table~\ref{tab:mf_bp_mlp_eff_ch4}.

\begin{table}[t!] 
    \centering
    \caption{Efficiency Comparison: MF vs. BP on Native MLP Architectures (Mean ± Std Dev over 3 runs).}
    \label{tab:mf_bp_mlp_eff_ch4}
    \resizebox{\textwidth}{!}{%
    \begin{tabular}{ccc c@{ ± }c c@{ ± }c c@{ ± }c c c}
        \toprule
        Dataset & Architecture & Algorithm & \multicolumn{2}{c}{Train Time (s)} & \multicolumn{2}{c}{Energy (Wh)} & \multicolumn{2}{c}{Est. CO2e (g)} & Peak Mem (MiB) & Total GFLOPs \\
        \midrule
        \multirow{2}{*}{MNIST} & \multirow{2}{*}{2 $\times$ 1000 MLP} & MF & \textbf{35.03} & \textbf{1.51} & \textbf{0.60} & \textbf{0.03} & \textbf{1.15} & \textbf{0.08} & 934 & \textbf{0.00359} \\
        & & BP  & 39.84 & 0.20 & 0.69 & 0.00 & 1.30 & 0.04 & \textbf{926} & 0.01077 \\
        \midrule
        \multirow{2}{*}{F-MNIST} & \multirow{2}{*}{2 $\times$ 1000 MLP} & MF  & 52.08 & 0.43 & 0.86 & 0.03 & 1.64 & 0.34 & 934 & \textbf{0.00359} \\
        & & BP  & \textbf{44.06} & \textbf{4.45} & \textbf{0.79} & \textbf{0.11} & \textbf{1.62} & \textbf{0.57} & \textbf{926} & 0.01077 \\
        \midrule
        \multirow{2}{*}{CIFAR-10} & \multirow{2}{*}{3 $\times$ 2000 MLP} & MF & \textbf{177.70} & \textbf{1.62} & \textbf{3.17} & \textbf{0.06} & \textbf{6.70} & \textbf{0.71} & \textbf{1120} & \textbf{0.02833} \\
        & & BP & 268.45 & 11.47 & 5.35 & 0.22 & 11.03 & 0.45 & 1184 & 0.08499 \\
        \midrule
        \multirow{2}{*}{CIFAR-100} & \multirow{2}{*}{3 $\times$ 2000 MLP} & MF & \textbf{110.36} & \textbf{1.33} & \textbf{2.02} & \textbf{0.03} & 3.46 & 0.35 & \textbf{1142} & \textbf{0.02869} \\
        & & BP & 111.78 & 12.44 & 2.30 & 0.26 & \textbf{3.29} & \textbf{0.48} & 1192 & 0.08607 \\
        \bottomrule
    \end{tabular}%
    }
    \caption*{\footnotesize Total GFLOPs: Estimated per sample ($F_{fwd}$ for MF, $F_{BP\_update}$ for BP). Best efficiency results in each group are in \textbf{bold} (lower is better).}
\end{table}

\begin{figure}[htbp]
    \centering
    \begin{subfigure}{1.0\textwidth}
        \centering
        \includegraphics[width=\linewidth]{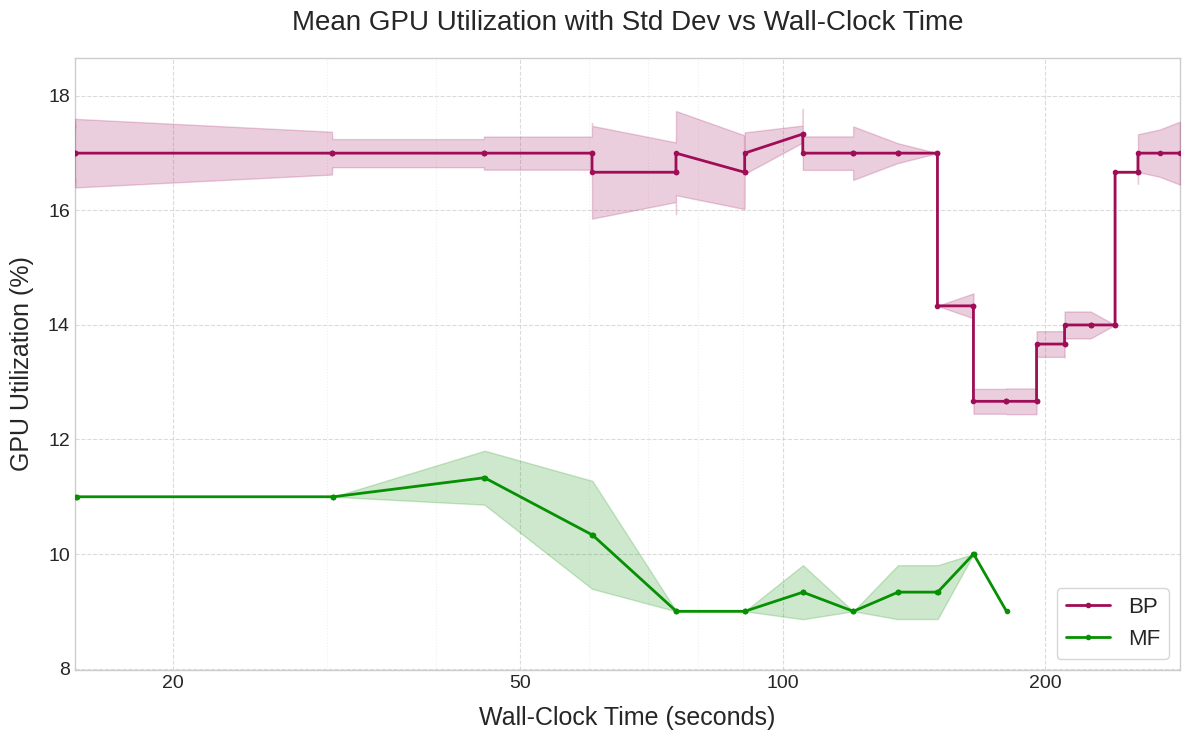}
        \caption{}
        \label{fig:mf_bp_gpu_util}
    \end{subfigure}
    \hfill
    \begin{subfigure}{1.0\textwidth}
        \centering
        \includegraphics[width=\linewidth]{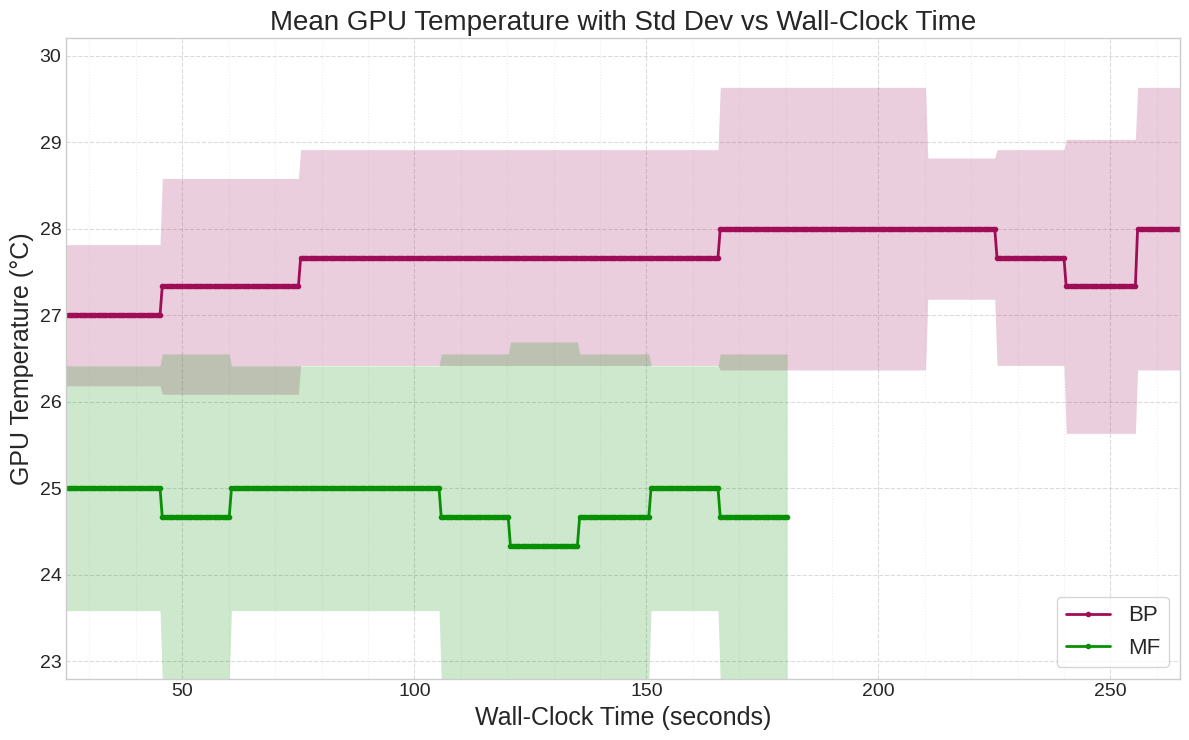} 
        \caption{}
        \label{fig:mf_bp_gpu_temp}
    \end{subfigure}
    \caption{Hardware-level Efficiency Indicators for MF vs. BP Training on CIFAR-10 3 $\times$ 2000 MLP.}
    \label{fig:mf_bp_hardware_indicators}
\end{figure}

\pagebreak
This superior efficiency is a direct consequence of its computationally lighter design. Hardware-level monitoring reveals that MF operates at a significantly lower computational intensity (9 to 11\% GPU utilization) compared to BP (16 to 17\%), a direct result of eliminating the backward pass, as seen in Figure~\ref{fig:mf_bp_gpu_util}. This reduced computational load leads to a reduced thermal profile, which is confirmed by the GPU temperature readings in Figure~\ref{fig:mf_bp_gpu_temp}. 

Although the efficiency gains in time and energy are profound, the analysis of peak memory consumption presents a more nuanced picture. Contrary to the common expectation that BP-free methods yield large memory savings, empirical evidence shows that the MF's advantage in this domain is modest. It realizes maximum memory reductions of approximately 4 to 5\% on the larger MLPs, while incurring a slight increase on the smaller ones (Table~\ref{tab:mf_bp_mlp_eff_ch4}).

\begin{figure}[htbp]
    \centering
    \includegraphics[width=0.99\textwidth]{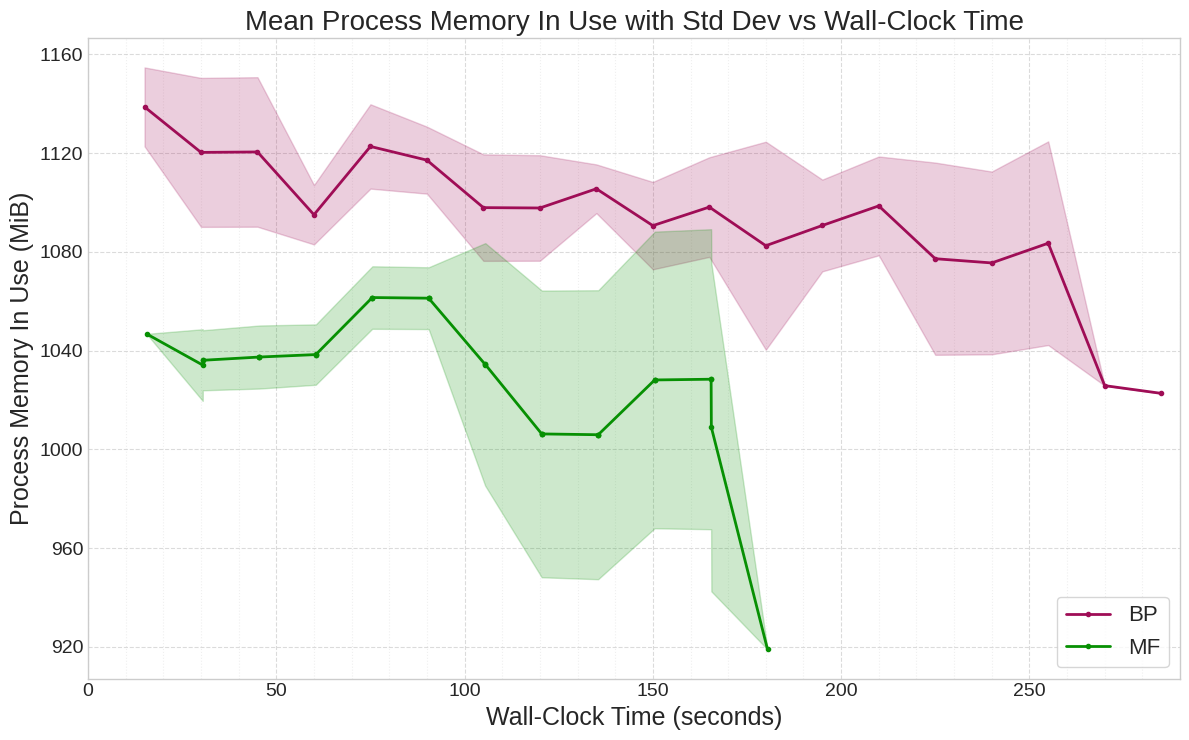}
    \caption{Mean Process Memory In Use for MF vs. BP on CIFAR-10 3 $\times$ 2000 MLP.}
    \label{fig:mf_bp_proc_mem}
\end{figure}

\pagebreak
This important finding highlights that the theoretical memory benefit of avoiding intermediate activation storage is partially offset by the practical memory overhead of MF's unique components: the learnable projection matrix ($\mathbf{M}_i$) associated with each layer and the memory required for its corresponding optimizer states. This result qualifies generalized claims and underscores the necessity of empirical measurement to develop a complete understanding of an algorithm's resource profile. Furthermore, the dynamic monitoring plots serve as powerful, independent signatures that validate MF's unique training mechanism is operating as designed. The dynamic memory plot in Figure~\ref{fig:mf_bp_proc_mem} provides a clear visual narrative, where the distinct step-like increase in MF's memory usage aligns perfectly with the sequential initiation of each layer's training. 

\subsection{Cross Algorithm Comparative Synthesis}
\label{sec:exp_cross_algo_summary_ch4}

To synthesize the empirical findings detailed in the preceding sections, this analysis provides a consolidated overview of the performance and efficiency results across all evaluated algorithms. Table~\ref{tab:cross_algo_summary_results_ch4_updated} quantifies the relative differences between each alternative training method and its corresponding fair backpropagation baseline, calculated from the mean results of three experimental runs. This comparative summary enables a direct assessment of the trade-offs inherent in each approach when implemented on its native architecture.

\begin{table}[t!]
    \centering
    \sisetup{
        table-format = +4.2,      
        retain-explicit-plus,     
        table-align-text-post = false 
    }
    \caption{Relative Performance and Efficiency Summary vs. Fair BP Baselines (Mean over 3 runs).}
    \label{tab:cross_algo_summary_results_ch4_updated}
    \resizebox{\textwidth}{!}{%
 \begin{tabular}{>{\centering\arraybackslash}m{3.2cm} l >{\centering\arraybackslash}m{2.8cm} S S S S}
        \toprule
        Algorithm Variant & Dataset & Architecture & {$\Delta$ Acc. (\%)} & {$\Delta$ Time (\%)} & {$\Delta$ Energy (\%)} & {$\Delta$ Mem. (\%)} \\
        \midrule
        FF-AdamW & F-MNIST & MLP 4 $\times$ 2000 & +0.75 & +1233.26 & +862.35 & +1.88 \\
        FF-AdamW & MNIST & MLP 3 $\times$ 1000 & +0.24 & +342.11 & +339.95 & +0.85 \\
        FF-SGD & MNIST & MLP 3 $\times$ 1000 & +0.14 & +429.69 & +419.67 & +1.06 \\
        FF-AdamW & MNIST & MLP 4 $\times$ 2000 & -0.01 & +305.08 & +236.88 & +1.88 \\
        \midrule
        CaFo-Rand-CE & MNIST & CNN 3-block& -0.32 & +88.48 & +32.65 & -6.67 \\
        CaFo-DFA-CE & MNIST & CNN 3-block& +0.08 & +124.82 & +81.83 & +1.11 \\
        \cmidrule(lr){2-7}
        CaFo-Rand-CE & F-MNIST & CNN 3-block& +1.11 & +205.00 & +115.99 & -6.67 \\
        CaFo-DFA-CE & F-MNIST & CNN 3-block& \bfseries +2.47 & +206.51 & +201.43 & +1.11 \\
        \cmidrule(lr){2-7}
        CaFo-Rand-CE  & CIFAR-10 & CNN 3-block& -13.23 & -2.96 & -19.24 & \bfseries -8.98 \\
        CaFo-DFA-CE & CIFAR-10 & CNN 3-block& -1.72 & +287.17 & +301.94 & +1.26 \\
        \cmidrule(lr){2-7}
        CaFo-Rand-CE  & CIFAR-100 & CNN 3-block& -11.44 & +246.27 & +188.87 & -5.02 \\
        CaFo-DFA-CE & CIFAR-100 & CNN 3-block& -4.43 & +557.19 & +576.82 & +2.51 \\
        \midrule
        MF & MNIST & MLP 2 $\times$ 1000 & +0.09 & -12.07 & -13.34 & +0.86 \\
        MF  & F-MNIST & MLP 2 $\times$ 1000 & +0.51 & +18.20 & +9.90 & +0.86 \\
        MF  & CIFAR-10 & MLP 3 $\times$ 2000 & +1.21 & \bfseries -33.81 & \bfseries -40.78 & -5.41 \\
        MF  & CIFAR-100 & MLP 3 $\times$ 2000 & +0.37 & -1.28 & -12.48 & -4.19 \\
        \bottomrule
    \end{tabular}%
    }
    \caption*{\footnotesize The metric $\Delta$ represents the difference between the alternative algorithm and its fair BP baseline. A positive $\Delta$Acc. is better. Negative values for other metrics are better. The \textbf{bold} values highlight the best in class improvement for each metric across all experiments and algorithms.}
\end{table}

The results consolidated in Table~\ref{tab:cross_algo_summary_results_ch4_updated} crystallize the main findings of this chapter. The Forward-Forward algorithm proved to be prohibitively inefficient, despite validating the foundational premise of BP-free learning by achieving an accuracy comparable to that of backpropagation on MLPs. Its training time and energy costs were consistently and substantially higher, without realizing any peak memory advantage in these experiments. The Cascaded-Forward algorithm, evaluated on its native CNNs, revealed a nuanced performance profile contingent on its implementation variant. The CaFo-Rand-CE variant demonstrated modest efficiency gains, including a notable 19.2\% energy reduction and a 9.0\% memory reduction on the CIFAR-10 task, but at the cost of a significant 13.23 percentage point decrease in accuracy.

In contrast, the CaFo-DFA-CE variant substantially improved accuracy, even surpassing BP on less complex datasets, yet the resource-intensive DFA pre-training phase led to profoundly greater time and energy consumption. Lastly, the Mono-Forward algorithm, when evaluated on its native MLP architectures, consistently presented the most compelling profile among the investigated alternatives. Not only did it achieve accuracy parity or superiority over its BP baselines, but it also delivered substantial reductions in training time and energy consumption, particularly on more complex datasets. The peak memory savings, while more modest, were consistent for the larger MLP architectures.

In aggregate, the experimental findings of this chapter chart an evolutionary progression from a conceptually significant but practically inefficient algorithm (FF), through a structured method defined by acute trade-offs (CaFo), to a highly promising algorithm (MF) that, within the domain of MLP architectures, effectively synthesizes superior accuracy with superior training efficiency. This empirical evidence provides the foundation for the critical analysis of algorithmic mechanisms and practical implications presented in the subsequent chapter.
\section{Discussion}
\label{chap:discussion}

This chapter synthesizes the experimental results from Chapter~\ref{chap:experiments}, critically evaluating the performance and efficiency profiles of the alternative algorithms against their respective BP baselines. The analysis focuses on interpreting the observed trade-offs, positioning the findings within the existing literature, and assessing the practical implications of these alternative training paradigms, with a particular focus on energy efficiency. The evolutionary narrative of BP-free methods, progressing from the FF's foundational demonstration through CaFo's structured design to the emergence of MF as a highly efficient and performant alternative, provides the guiding framework for this discussion.

\subsection{Interpretation of Key Findings}
\label{sec:discussion_interpretation}

The empirical results presented in Chapter~\ref{chap:experiments} provide a nuanced perspective on the capabilities of BP-free algorithms, tracing an evolutionary path toward the dual objectives of competitive accuracy and superior efficiency.

Geoffrey Hinton's FF algorithm established a crucial proof of concept by validating that deep networks can attain competitive accuracy without a backward pass (Table~\ref{tab:ff_bp_mlp_perf_results_ch4}). However, this validation was performed at a prohibitive efficiency cost. As illustrated in Figure~\ref{fig:ch4_ff_bp_mnist_mlp_4x2000_conv_time}, FF's convergence process is markedly slower and more volatile than that of backpropagation, culminating in significantly extended training durations and elevated energy consumption. The analysis at the hardware level (Figure~\ref{fig:ch4_ff_bp_mnist_mlp_4x2000_resource_util}) critically revealed that FF's computational patterns result in suboptimal GPU utilization and that its presumed memory advantages did not materialize for the MLP architectures tested. This positions FF as a \textbf{conceptually significant yet practically unviable baseline}, thereby motivating the investigation of more sophisticated methods.

The CaFo algorithm signifies a structured advancement, designed to deliver a more direct supervised signal via its block-wise predictor architecture. The experimental findings revealed a clear trade-off between its two evaluated variants. CaFo-Rand-CE, which uses fixed random blocks, offered consistent, though modest, peak memory savings, but incurred a substantial accuracy penalty on complex datasets (Table~\ref{tab:cafo_bp_cnn_perf_results_ch4_reordered}). This result highlights that effective learning cannot be achieved simply by adding local classifiers to random feature projections. In contrast, CaFo-DFA-CE substantially improved accuracy by pre-training its blocks with DFA, approaching BP's performance. However, this success was accompanied by a considerable computational burden, as the DFA pre-training phase led to profoundly higher energy and time consumption (Table~\ref{tab:cafo_bp_cnn_eff_results_ch4_reordered_eff}). CaFo consequently presents a challenging decision: one must either accept a significant performance deficit for a niche efficiency gain or incur a heavy computational cost to achieve competitive accuracy.

The MF algorithm, the most advanced method investigated, emerges from these experiments as a particularly compelling alternative for its native MLP architectures. As presented in Table~\ref{tab:mf_bp_mlp_perf_ch4}, MF \textbf{consistently achieved or exceeded the accuracy of rigorously optimized BP baselines}. This enhanced generalization performance is attributed to its sequence of local optimizations converging to a more advantageous final validation loss compared to BP's global optimization strategy (Figure~\ref{fig:mf_bp_cifar10_mlp_conv_curves}). This superior performance coincided with profound efficiency improvements. On the CIFAR-10 MLP task, MF proved to be approximately 34\% faster and consumed approximately 41\% less energy than its BP counterpart (Table~\ref{tab:mf_bp_mlp_eff_ch4}). This efficiency is a direct consequence of its computationally lighter design, which eliminates the backward pass. MF thus effectively navigates the trade-off between accuracy and efficiency, \textbf{substantially fulfilling the promise of a practical, high-performance, and sustainable BP-free algorithm for MLPs}.

A critical methodological element supporting these interpretations is the consistent application of systematic hyperparameter Optuna optimization and validation-based early stopping. This protocol ensured that every algorithm was evaluated at its optimal performance point, facilitating a fair comparison of their intrinsic characteristics.

\subsection{Algorithm and Architecture Strengths and Weaknesses}
\label{sec:discussion_strengths_weaknesses}

Each algorithm demonstrated distinct strengths and weaknesses that were frequently related to its unique learning mechanism and its interaction with the native network architecture. The Forward-Forward (FF) algorithm's primary strength lies in its conceptual contribution: It fundamentally established that BP-free local learning can achieve competitive accuracy. However, its weaknesses are profound from a practical point of view, including a high inefficiency in both time and energy.

The Cascaded-Forward (CaFo) algorithm presents a more complex set of trade-offs. Its main weakness stems from a crucial principle revealed by these results: \textbf{the quality of feature representations is paramount in BP-free learning}. The mere attachment of local predictors, as in the Rand-CE variant, is insufficient for complex tasks. Conversely, while the DFA-CE variant produces effective features, its pre-training stage is exceptionally resource-intensive, negating any overall energy efficiency.

Mono-Forward (MF) on MLPs demonstrated the most compelling balance of strengths. It consistently achieved accuracy on par with or superior to fair BP baselines, coupled with significant reductions in training time and energy consumption. However, a critical and counterintuitive weakness emerged from the analysis: \textbf{The theoretical memory savings of BP-free learning did not fully materialize}. The peak memory advantage was modest or non-existent on the tested MLPs. This empirically demonstrates that the practical memory overhead of MF's unique components, specifically the learnable projection matrices and their optimizer states, \textbf{can partially or fully counteract the gains} from forgoing activation storage, challenging a common assumption about BP-free methods.

Finally, backpropagation (BP) serves as a benchmark for a highly mature, robust, and generally efficient algorithm, benefiting from decades of co-optimization. Its well-known weakness is the inherent memory and computational cost of the backward pass. This analysis highlights that the optimal approach is context-dependent, with MF exhibiting a particularly strong profile for applications based on MLPs.

\subsection{Practical Implications}
\label{sec:discussion_implications}

The results of this paper carry several practical implications for the development and deployment of deep learning models, particularly concerning sustainability. The most direct consequence is the potential for significant energy savings and a reduced environmental impact. By delivering substantial reductions in training time and energy consumption while enhancing accuracy, the MF algorithm \textbf{presents a concrete path toward more sustainable AI}, translating directly into a lower estimated CO2e. These efficiency gains also make such algorithms highly advantageous for deployment on edge devices or in other resource-constrained environments. The emergence of effective BP-free algorithms like MF \textbf{could also inspire new hardware architectures} optimized for their specific computational patterns, potentially altering trade-offs in future chip designs. Ultimately, this research provides empirical data to inform algorithm selection. For tasks based on MLPs, \textbf{MF appears to be a better choice than BP when both accuracy and efficiency are primary considerations}, suggesting that this direct local loss paradigm is a powerful direction for future algorithm development.

\pagebreak

\subsection{Theoretical Insights}
\label{sec:discussion_theoretical}

This comparative investigation also yields several theoretical insights into the mechanisms of deep learning. The results robustly demonstrate that \textbf{effective high-performance learning can be realized using purely local rules}, a foundational premise established by FF and compellingly validated by MF. The success of MF's mechanism suggests that complex credit assignment can be managed without global error propagation. By eschewing the precise symmetric weight transport of BP, these algorithms align more closely with the principles of biological learning. A key related insight is the strong coupling between the algorithm and its architecture, suggesting that \textbf{future progress in BP-free learning will likely depend on the co-design of both the learning rule and the network primitives} upon which it operates. Finally, MF's ability to converge to a lower final validation loss than BP on the CIFAR-10 MLP task (Figure~\ref{fig:mf_bp_cifar10_mlp_conv_curves}) offers a fascinating insight into optimization landscapes. It suggests that a sequence of greedy local optimizations may guide a model to a more advantageous region of the parameter space, \textbf{challenging the conventional wisdom that only global optimization can locate the best solutions}.

\subsection{Limitations and Critical Evaluation}
\label{sec:discussion_limitations}

Although this paper provides a rigorous comparison under controlled conditions, several limitations must be acknowledged. First, the author intentionally focused the experimental scope on the \textit{native} architectures for each algorithm. Consequently, the \textbf{strong performance of MF on MLPs does not automatically guarantee similar success on different architectures}, such as Transformers or deeper CNNs, without further research.

A second set of limitations arises from algorithm-specific nuances. The inefficiency of FF makes its current formulation impractical. For CaFo, the DFA-CE variant proved too resource-intensive to be "efficient", while the Rand-CE variant's accuracy remains a significant weakness. Most importantly, although highly successful on MLPs, \textbf{MF's peak memory savings were modest}, a critical nuance highlighting that the overhead of its unique components can counteract a significant portion of the gains from eliminating the backward pass.

Furthermore, while systematic hyperparameter optimization was conducted, such searches are not exhaustive. The findings are also tied to a specific hardware and software stack, and the estimated CO2e depends on the local grid's carbon intensity. Finally, this paper focuses primarily on training efficiency; a detailed analysis of inference performance was beyond its primary scope. Despite these limitations, this work contributes a detailed and fair empirical comparison to the field of BP-free deep learning, particularly by highlighting the potential of MF as a practical, high-performance, and energy-efficient algorithm for applications based on MLPs.
\section{Conclusion}
\label{chap:conclusion}

This paper presented a rigorous investigation into the performance and energy efficiency of three BP-free training algorithms: FF, CaFo, and MF, benchmarked against standard backpropagation. Through a framework of fair comparisons on native architectures, incorporating direct hardware measurements, systematic hyperparameter optimization, and validation-based early stopping, this work clarifies the practical viability of these alternatives for more sustainable deep learning. The evolutionary trajectory from FF's foundational proof of concept, through CaFo's structured design, to MF's emergence as a highly efficient and performant method for MLPs highlights the significant potential and rapid advancement within this domain.

\subsection{Key Findings and Answers to Research Questions}
\label{sec:conclusion_findings}

The experimental results provide definitive answers to the research questions posed at the beginning of this work. Regarding classification performance, the MF algorithm, in its native MLP architectures, \textbf{robustly and consistently matched or surpassed the accuracy of its BP baseline}. Its convergence to a lower final validation loss indicates that its layer-wise optimization strategy identified a more effective solution. In contrast, FF achieved comparable accuracy but with markedly slower and more volatile convergence, while the CaFo-Rand-CE variant revealed a clear performance deficit on complex datasets. The CaFo-DFA-CE variant dramatically narrowed this gap, but at a significant computational cost during pre-training.

In terms of direct energy consumption, the MF algorithm yielded significant savings, consuming approximately 41\% less energy than BP on the CIFAR-10 MLP task while achieving superior accuracy. This is in stark contrast to the case for FF, which consumed substantially more energy because of its inefficiency, and CaFo-DFA-CE, which was also more energy-intensive. This research also \textbf{empirically refutes the generalized assumption that all BP-free methods are inherently more memory efficient}. No algorithm yielded dramatic memory savings; for MF on larger MLPs, the peak memory advantage was modest (4\% to 5\%), as the overhead for its additional projection matrices was found to partially offset the theoretical gains.

These results clarify the different energy-performance trade-offs. FF offers an unfavorable trade-off of competitive accuracy for prohibitive inefficiency. CaFo presents a stark choice between the Rand-CE variant's modest efficiency gains for a significant accuracy drop, and the DFA-CE variant's competitive accuracy for a substantial efficiency cost. Crucially, MF delivered the most advantageous trade-off, achieving \textbf{accuracy parity or superiority with simultaneous and significant improvements in training time and energy efficiency}. It also demonstrated effective scalability in both performance and efficiency across the tested MLP and dataset complexities. In summary, this research established FF as a foundational proof of concept, identified CaFo as a structured method defined by clear trade-offs, and distinguished MF as a highly promising algorithm that, for MLP architectures, fulfills the objectives of achieving both BP-competitive accuracy and superior training efficiency.

\subsection{Summary of Contributions}
\label{sec:conclusion_contributions}

This paper makes several key contributions to the study of energy-efficient deep learning. First, it established a \textbf{rigorous and fair benchmarking framework} by replicating the native architectures for FF, CaFo, and MF, creating identical BP baselines, and applying systematic, validation-based optimization to all methods. This work presented one of the first \textbf{extensive independent evaluations of the recently proposed MF algorithm} on its native MLP architectures, robustly demonstrating its ability to improve both accuracy and efficiency. Through detailed profiling centered on \textbf{direct measurements of GPU energy consumption via NVML}, this research provided critical insights into the inefficient hardware utilization of FF and the nuanced memory characteristics of MF.
\pagebreak

This research systematically detailed the trade-offs for each algorithm, clarifying that MF (on MLPs) provided the most favorable balance. Finally, to promote reproducibility and facilitate future research, all developed code, structured configurations, experimental protocols, and measurement tools were organized into an open research infrastructure and made publicly available.

\subsection{Future Research Directions}
\label{sec:conclusion_future_work}

The findings of this paper suggest several promising avenues for future research. A primary direction is architectural exploration, particularly investigating \textbf{MF's adaptability and performance on more complex architectures} like CNNs and Transformers, which will require adapting its projection matrix mechanism. For CaFo, developing more computationally efficient, BP-free methods for training its neural blocks is essential. Further research should also evaluate the most promising algorithms on larger-scale datasets like ImageNet to fully ascertain their scalability. Other important frontiers include developing custom hyperparameter pruners for layer-wise training, deepening the theoretical understanding of how MF's local optimization can achieve superior global solutions, and pursuing \textbf{hardware co-design for BP-free algorithms} to unlock additional energy savings. Extending these methods beyond classification to domains like generative modeling and reinforcement learning, as well as investigating their robustness and uncertainty calibration, will be crucial for their broader adoption.

\subsection{Broader Implications}
\label{sec:conclusion_implications}

The research detailed in this paper has broad implications for the field. As artificial intelligence becomes increasingly prevalent, the energy consumption of these systems has become a significant social concern. The development and adoption of algorithms such as MF, which can significantly reduce the footprint of training energy without a performance penalty, represent \textbf{critical steps toward a more sustainable future for AI}. Lowering energy requirements can also \textbf{democratize AI development}, making it more accessible to researchers with limited computational resources, and is paramount for enabling powerful on-device learning under strict power constraints. The biological inspiration underlying these algorithms fosters a symbiotic relationship between AI and neuroscience, where brain-inspired computation can inform more efficient AI. Ultimately, this research contributes to a paradigm shift in which efficiency is treated as a primary design criterion, on par with accuracy. The journey beyond backpropagation is not just about devising novel algorithms; it is about \textbf{shaping a more efficient, accessible, and responsible future for artificial intelligence}. 

%
%

\subsection*{Acknowledgments}
The authors would like to thank Professor Witold Dzwinel for his invaluable guidance and expert supervision. This work was made possible by the computational resources provided by ACK Cyfronet AGH on the Athena cluster. The research was supported by the PLGrid Infrastructure.


%
%

\pagebreak 
\appendix

\section{Hyperparameter Tuning Details}
\label{app:hyper_tuning}
This appendix provides a comprehensive account of the hyperparameter settings established for each algorithm in this thesis. To ensure a rigorous and fair comparison, systematic hyperparameter optimization was conducted for all training algorithms. This included Backpropagation (BP), Forward-Forward (FF), Cascaded-Forward (CaFo), and Mono-Forward (MF), each tested across their respective experimental configurations. The process was facilitated by the Optuna framework \cite{akiba2019optuna}, a tool the author employed to methodically search for optimal parameters.

The general optimization strategy utilized a Tree-structured Parzen Estimator (TPE) sampler \cite{bergstra2011algorithms}. For the BP baselines, a Median pruner was implemented to terminate unpromising trials based on intermediate validation performance. Conversely, pruning was disabled for the alternative algorithms (FF, CaFo, MF), as their distinct, often layer-wise, training dynamics render intermediate metrics unreliable for predicting final outcomes. The primary objective for each tuning trial was the maximization of validation set accuracy, evaluated within a specified training budget. Upon completion of each search, the identified optimal hyperparameters were automatically propagated to the final experiment configuration files using dedicated utility scripts. The resulting optimized parameters, which were used for the final experimental runs detailed in Chapter \ref{chap:experiments}, are documented in the following sections.

\subsection{Backpropagation Baseline Tuning Parameters}
The AdamW optimizer \cite{loshchilov2017decoupled} was utilized for all BP baseline experiments. The optimization process was configured as follows:
\begin{itemize}
    \item \textbf{Search Space (Common):}
        \begin{itemize}
            \item Learning Rate (\texttt{lr}): Loguniform($10^{-5}$, $10^{-2}$)
            \item Weight Decay (\texttt{weight\_decay}): Loguniform($10^{-6}$, $10^{-3}$)
        \end{itemize}
    \item \textbf{Objective:} Maximize validation accuracy within the trial's training budget, which incorporated early stopping based on validation loss.
    \item \textbf{Sampler/Pruner:} TPE / Median.
\end{itemize}
The optimal hyperparameters identified for each BP baseline configuration were:
\begin{itemize}
    \item \textbf{BP Baseline for FF (Fashion-MNIST 4x2000 MLP):}
        \begin{itemize}
            \item Learning Rate: 0.0002040628171913876
            \item Weight Decay: 4.186646492145325e-06
        \end{itemize}
    \item \textbf{BP Baseline for CaFo (Fashion-MNIST 3-Block CNN):}
        \begin{itemize}
            \item Learning Rate: 0.000564093098959669
            \item Weight Decay: 8.597071465954117e-06
        \end{itemize}
     \item \textbf{BP Baseline for CaFo (CIFAR-10 3-Block CNN):}
        \begin{itemize}
            \item Learning Rate: 0.0067628739201322655
            \item Weight Decay: 4.7418070637518205e-06
        \end{itemize}
     \item \textbf{BP Baseline for CaFo (CIFAR-100 3-Block CNN):}
        \begin{itemize}
            \item Learning Rate: 0.0003201243830098061
            \item Weight Decay: 5.880350745731698e-05
        \end{itemize}
     \item \textbf{BP Baseline for MF (Fashion-MNIST 2x1000 MLP):}
        \begin{itemize}
            \item Learning Rate: 0.00021316290979390737
            \item Weight Decay: 2.7122805865833542e-05
        \end{itemize}
    \item \textbf{BP Baseline for MF (CIFAR-10 3x2000 MLP):}
            \begin{itemize}
                \item Learning Rate: 8.11587582031004e-05
                \item Weight Decay: 2.0847447927718932e-05
            \end{itemize}
    \item \textbf{BP Baseline for MF (CIFAR-100 3x2000 MLP):}
            \begin{itemize}
                \item Learning Rate: 0.00010362161015212055
                \item Weight Decay: 9.105881789069153e-06
            \end{itemize}
    \item \textbf{BP Baseline for FF (MNIST 3x1000 MLP):}
        \begin{itemize}
            \item Learning Rate: 0.0001329291894316216
            \item Weight Decay: 0.0007114476009343421
        \end{itemize}
    \item \textbf{BP Baseline for FF (MNIST 4x2000 MLP):}
        \begin{itemize}
            \item Learning Rate: 0.0001329291894316216
            \item Weight Decay: 0.0007114476009343421
        \end{itemize}
    \item \textbf{BP Baseline for CaFo (MNIST 3-Block CNN):}
        \begin{itemize}
            \item Learning Rate: 0.001570297088405539
            \item Weight Decay: 6.251373574521755e-05
        \end{itemize}
    \item \textbf{BP Baseline for MF (MNIST 2x1000 MLP):}
        \begin{itemize}
            \item Learning Rate: 0.00019762189340280086
            \item Weight Decay: 7.4763120622522945e-06
        \end{itemize}
\end{itemize}

\subsection{Alternative Algorithm Tuning Parameters}
The alternative algorithms were tuned using their respective dedicated Optuna objective functions.
\begin{itemize}
    \item \textbf{Objective:} Maximize final validation accuracy at the end of the algorithm's specific training procedure.
    \item \textbf{Sampler/Pruner:} TPE / None.
\end{itemize}

\textbf{Forward-Forward (FF):}
\begin{itemize}
    \item \textbf{Search Space (Common):}
        \begin{itemize}
            \item FF Layer Learning Rate (\texttt{ff\_learning\_rate}): Loguniform ($2 \times 10^{-4}$, $5 \times 10^{-3}$)
            \item FF Layer Weight Decay (\texttt{ff\_weight\_decay}): Loguniform ($10^{-4}$, $10^{-3}$)
            \item Downstream Classifier Learning Rate (\texttt{downstream\_learning\_rate}): Loguniform ($2 \times 10^{-3}$, $5 \times 10^{-2}$)
            \item Downstream Classifier Weight Decay (\texttt{downstream\_weight\_decay}): Loguniform ($10^{-3}$, $10^{-2}$)
        \end{itemize}
    \item \textbf{Best Hyperparameters Found:}
        \begin{itemize}
            \item FF (Fashion-MNIST 4x2000 MLP, AdamW):
                \begin{itemize}
                    \item \texttt{ff\_learning\_rate}: 0.0003722230492626773
                    \item \texttt{ff\_weight\_decay}: 0.0003583569744251919
                    \item \texttt{downstream\_learning\_rate}: 0.011120232890144957
                    \item \texttt{downstream\_weight\_decay}: 0.0051200359925764015
                \end{itemize}
             \item FF (MNIST 3x1000 MLP, SGD):
                 \begin{itemize}
                     \item \texttt{ff\_learning\_rate}: 0.0019075980272792064
                     \item \texttt{ff\_weight\_decay}: 0.0005229088821652552
                     \item \texttt{downstream\_learning\_rate}: 0.03234998392840941
                     \item \texttt{downstream\_weight\_decay}: 0.0047668064095875455
                     \item \texttt{ff\_momentum}: 0.9
                     \item \texttt{downstream\_momentum}: 0.9
                 \end{itemize}
             \item FF (MNIST 3x1000 MLP, AdamW):
                 \begin{itemize}
                     \item \texttt{ff\_learning\_rate}: 0.0004759134394296565
                     \item \texttt{ff\_weight\_decay}: 0.0004169096909947932
                     \item \texttt{downstream\_learning\_rate}: 0.010727571797342173
                     \item \texttt{downstream\_weight\_decay}: 0.0061527185182837065
                 \end{itemize}
             \item FF (MNIST 4x2000 MLP, AdamW):
                 \begin{itemize}
                     \item \texttt{ff\_learning\_rate}: 0.0005253511034001832
                     \item \texttt{ff\_weight\_decay}: 0.0004227844165066995
                     \item \texttt{downstream\_learning\_rate}: 0.0027075510107757122
                     \item \texttt{downstream\_weight\_decay}: 0.0035730744916099627
                 \end{itemize}
        \end{itemize}
     \item \textbf{Fixed Parameters:} Key parameters were held constant during the tuning of the FF algorithm. These included the optimizer type (AdamW or SGD), a goodness threshold ($\theta=2.0$), and peer normalization settings (factor=0.03, momentum=0.9). All trials were conducted for a maximum of 100 epochs, with early stopping triggered by the \texttt{FF\_Hinton/Val\_Acc\_Epoch} metric, using a patience of 20 epochs and a minimum delta of 0.01.
\end{itemize}

\textbf{Cascaded-Forward (CaFo):}
\begin{itemize}
    \item \textbf{Search Space (Common):}
        \begin{itemize}
            \item Predictor Learning Rate (\texttt{predictor\_lr}): Loguniform ($2 \times 10^{-4}$, $5 \times 10^{-3}$)
            \item Predictor Weight Decay (\texttt{predictor\_weight\_decay}): Loguniform ($10^{-7}$, $10^{-4}$)
            \item Epochs Per Predictor (\texttt{num\_epochs\_per\_block}): IntUniform (range dependent on dataset)
            \item \textit{For CaFo-DFA variant (\texttt{train\_blocks}=true):}
                \begin{itemize}
                    \item Block Learning Rate (\texttt{block\_lr}): Loguniform ($2 \times 10^{-5}$, $5 \times 10^{-4}$)
                    \item Block Weight Decay (\texttt{block\_weight\_decay}): Loguniform ($10^{-7}$, $10^{-4}$)
                \end{itemize}
        \end{itemize}
    \item \textbf{Best Hyperparameters Found:}
        \begin{itemize}
            \item CaFo-Rand-CE (Fashion-MNIST 3-Block CNN):
                \begin{itemize}
                    \item \texttt{predictor\_lr}: 0.0013737845955327983
                    \item \texttt{predictor\_weight\_decay}: 2.937538457632828e-07
                    \item \texttt{num\_epochs\_per\_block}: 145 (Maximum, with early stopping patience of 8)
                \end{itemize}
            \item CaFo-DFA-CE (Fashion-MNIST 3-Block CNN):
                 \begin{itemize}
                     \item \texttt{block\_lr}: 0.0001384690524742107
                     \item \texttt{block\_weight\_decay}: 1.3311216080736884e-05
                     \item \texttt{predictor\_lr}: 0.0003304463634024569
                     \item \texttt{predictor\_weight\_decay}: 3.967605077052991e-05
                     \item \texttt{num\_epochs\_per\_block}: 85 (Maximum, with early stopping patience of 6)
                 \end{itemize}
             \item CaFo-Rand-CE (CIFAR-10 3-Block CNN):
                 \begin{itemize}
                     \item \texttt{predictor\_lr}: 0.00024111676098423975
                     \item \texttt{predictor\_weight\_decay}: 6.3583588566762514e-06
                     \item \texttt{num\_epochs\_per\_block}: 733 (Maximum, with early stopping patience of 8)
                 \end{itemize}
             \item CaFo-DFA-CE (CIFAR-10 3-Block CNN):
                 \begin{itemize}
                     \item \texttt{block\_lr}: 0.0001373784595532798
                     \item \texttt{block\_weight\_decay}: 2.938027938703532e-07
                     \item \texttt{predictor\_lr}: 0.0006677511008261821
                     \item \texttt{predictor\_weight\_decay}: 1.5702970884055385e-05
                     \item \texttt{num\_epochs\_per\_block}: 147 (Maximum, with early stopping patience of 8)
                 \end{itemize}
             \item CaFo-Rand-CE (CIFAR-100 3-Block CNN):
                 \begin{itemize}
                     \item \texttt{predictor\_lr}: 0.00024111676098423975
                     \item \texttt{predictor\_weight\_decay}: 6.3583588566762514e-06
                     \item \texttt{num\_epochs\_per\_block}: 1220 (Maximum, with early stopping patience of 10)
                 \end{itemize}
             \item CaFo-DFA-CE (CIFAR-100 3-Block CNN):
                 \begin{itemize}
                     \item \texttt{block\_lr}: 0.0001373784595532798
                     \item \texttt{block\_weight\_decay}: 2.938027938703532e-07
                     \item \texttt{predictor\_lr}: 0.0006677511008261821
                     \item \texttt{predictor\_weight\_decay}: 1.5702970884055385e-05
                     \item \texttt{num\_epochs\_per\_block}: 1271 (Maximum, with early stopping patience of 10)
                 \end{itemize}
            \item CaFo-Rand-CE (MNIST 3-Block CNN):
                \begin{itemize}
                    \item \texttt{predictor\_lr}: 0.00042136234742798395
                    \item \texttt{predictor\_weight\_decay}: 1.6585525961391826e-06
                    \item \texttt{num\_epochs\_per\_block}: 227 (Maximum, with early stopping patience of 8)
                \end{itemize}
            \item CaFo-DFA-CE (MNIST 3-Block CNN):
                 \begin{itemize}
                     \item \texttt{block\_lr}: 0.0004977965859721583
                     \item \texttt{block\_weight\_decay}: 9.594431243311431e-07
                     \item \texttt{predictor\_lr}: 0.00025391813833013565
                     \item \texttt{predictor\_weight\_decay}: 2.0840357948896564e-06
                     \item \texttt{num\_epochs\_per\_block}: 12 (Maximum, with early stopping patience of 5)
                 \end{itemize}
        \end{itemize}
    \item \textbf{Fixed Parameters:} Several parameters for the CaFo algorithm remained fixed. The block architecture was a three-block CNN, with each block comprising a 3x3 Convolution, a ReLU activation, a 2x2 Max-Pooling layer, and Batch Normalization, with output channels of 32, 128, and 512 respectively. The Adam optimizer and Cross-Entropy loss were consistently used for predictor training, and the final predictions were aggregated by summation.
\end{itemize}

\textbf{Mono-Forward (MF):}
\begin{itemize}
    \item \textbf{Search Space (Common):}
        \begin{itemize}
            \item Learning Rate (\texttt{lr}): Loguniform ($10^{-5}$, $10^{-2}$)
            \item Epochs Per Layer (\texttt{epochs\_per\_layer}): IntUniform (e.g., 5-30)
        \end{itemize}
    \item \textbf{Best Hyperparameters Found:}
        \begin{itemize}
            \item MF (MNIST 2x1000 MLP):
                \begin{itemize}
                    \item Learning Rate: 0.001570297088405539
                    \item Epochs Per Layer: 14 
                \end{itemize}
            \item MF (Fashion-MNIST 2x1000 MLP):
                \begin{itemize}
                    \item Learning Rate: 0.0003873086262136253
                    \item Epochs Per Layer: 12 
                \end{itemize}
            \item MF (CIFAR-10 3x2000 MLP):
                \begin{itemize}
                    \item Learning Rate: 0.0003241756767272183
                    \item Epochs Per Layer: 15 
                \end{itemize}
            \item MF (CIFAR-100 3x2000 MLP):
                \begin{itemize}
                    \item Learning Rate: 0.00017277890583771544
                    \item Epochs Per Layer: 8 
                \end{itemize}
        \end{itemize}
    \item \textbf{Fixed Parameters:} For the MF algorithm, several key parameters were held constant to ensure consistency. The MLP architecture was fixed as specified for each dataset: a 2x1000 MLP for MNIST and Fashion-MNIST, and a 3x2000 MLP for CIFAR-10 and CIFAR-100. Layer-wise training was conducted using the Adam optimizer with a weight decay value fixed at 0.0, minimizing a local Cross-Entropy loss at each layer. All layer weights and their corresponding projection matrices were initialized using the Kaiming Uniform method.
\end{itemize}

\pagebreak

\section{Computational Environment}
\label{app:environment}
The experiments presented in this thesis were conducted within a consistent and well-defined computational environment hosted on the Athena cluster. The specific hardware and software configurations are detailed below to ensure the reproducibility of the presented results.
\begin{itemize}
    \item \textbf{Operating System:} Rocky Linux 9.5 (Blue Onyx)
    \item \textbf{CPU Architecture:} x86\_64 (AMD EPYC 7742 @ 2.25 GHz)
    \item \textbf{GPU Architecture:} NVIDIA Ampere (A100-SXM4-40GB)
    \item \textbf{NVIDIA Driver Version:} 570.86.15
    \item \textbf{CUDA Toolkit Version:} 12.4.0
    \item \textbf{Deep Learning Framework:} PyTorch v2.4.0 (build `+cu121` linked against CUDA 12.1)
    \item \textbf{Python Version:} 3.10.4
    \item \textbf{Key Python Libraries:}
        \begin{multicols}{3}
            \begin{itemize}
                \item \code{torchvision}: 0.19.0
                \item \code{optuna}: 4.2.1
                \item \code{wandb}: 0.19.8
                \item \code{pynvml}: 12.0.0
                \item \code{codecarbon}: 3.0.1
                \item \code{numpy}: 2.2.4
                \item \code{PyYAML}: 6.0.2
                \item \code{scikit-learn}: 1.6.1
                \item \code{tqdm}: 4.67.1 
            \end{itemize}
        \end{multicols}
    \item \textbf{Profiling Tool for FLOPs:} PyTorch Profiler (\texttt{torch.profiler})
    \item \textbf{Energy and Resource Monitoring Tool:} NVIDIA Management Library (NVML) via the \code{pynvml} package
    \item \textbf{Carbon Footprint Estimation Tool:} CodeCarbon v3.0.1 \cite{codecarbon2021}
    \item \textbf{Job Scheduler:} SLURM 23.11.7
\end{itemize}

\section{Code Repository and Experimental Logs}
\label{app:code_repository}
To ensure the highest degree of reproducibility and transparency, the complete source code for all experiments, along with the unabridged output logs from every run, has been made publicly available. This repository contains the full implementations of the alternative algorithms and their backpropagation baselines, all scripts for data loading and experiment execution, and the raw log files for complete verifiability of the results. The repository is hosted on GitHub at the following address: \url{https://github.com/Przemyslaw11/BeyondBackpropagation}

Comprehensive instructions for setting up the environment and replicating the experiments are provided in the \code{README.md} file within the repository. For specific software versions and hardware details, the author refers the reader to the environmental specifications listed in Appendix \ref{app:environment}.

\end{document}